\documentclass{article}
\usepackage{graphicx}
\usepackage{subcaption}
\usepackage{wrapfig}
\usepackage{lipsum}
\usepackage{printlen}
\usepackage{amsmath}
\usepackage{amssymb}
\usepackage{bbm}
\usepackage{csquotes}
\usepackage[export]{adjustbox}
\usepackage{float}

\newcommand{\textbfplus}[1]{\vspace{0.5mm}\textbf{#1}\;\;}

\usepackage[preprint]{neurips_2026}

\usepackage[utf8]{inputenc}
\usepackage[T1]{fontenc}    
\usepackage{hyperref}       
\usepackage{cleveref}
\usepackage{url}         
\usepackage{booktabs}       
\usepackage{amsfonts}     
\usepackage{nicefrac}      
\usepackage{microtype}     
\usepackage{xcolor}         
\setcitestyle{numbers}
 
\title{Learning Through Noise: Why Subliminal Learning Works and When It Fails}

\author{
    Vincent C. Brockers$^{1,2,}$\footnotemark[2]\and 
    \textbf{Roman D. Ventzke}$^{1,2,}$\footnotemark[2]\and
    \textbf{Valentin Neuhaus}$^{2,1,}$\footnotemark[2]\and
    \textbf{Belén Hidalgo-Ogalde}$^{1,2}$ \qquad \textbf{Viola Priesemann}$^{1,2}$\\\\
    $^1$Max Planck Institute for Dynamics and Self-Organization, Göttingen \\
    $^2$Faculty of Physics, Institute for the Dynamics of Complex Systems, University of Göttingen\\
    \texttt{\{vincent.brockers, viola.priesemann\}@ds.mpg.de}
}

\begin{document}

\maketitle

\begin{abstract}

In the context of artificial neural networks, subliminal learning refers to the transfer of task-relevant knowledge or unintended biases from teacher to student models through distillation on task-unrelated input--output pairs. Prior explanations tie this effect to shared or closely matched teacher--student initialization. We show that a closely matched initialization is not necessary. Instead, subliminal learning is governed by compatible output heads. Using a controlled MNIST setting, we split outputs into an auxiliary head (for auxiliary, task-unrelated noise signals) and a class head (for classification) to demonstrate subliminal learning occurs---even when we randomly initialize hidden layers and remove layers, add new layers, or change the architecture (MLP-to-CNN). Compatible auxiliary heads enable transfer of a recoverable teacher signal, bringing the student's representations closer to the teacher’s. When the class heads remain compatible as well, students trained only on task-unrelated noise can approach, and in favorable regimes match, teacher-level task performance. Our setting enables us to develop a theory that explains the mechanism of subliminal learning and to derive upper bounds on when subliminal learning fails. Together, our results turn subliminal learning from a surprising transfer effect into a theoretically grounded mechanism with predictable limits.
\end{abstract}

\newpage
\section{Introduction} 
Broadly speaking, subliminal learning is the phenomenon of learning from signals/stimuli below the threshold of perception~\cite{watanabe2001perceptual, seitz2003subliminal}. 
Surprisingly, a similar phenomenon was recently observed in large language models (LLMs), when training a student network on task-unrelated random digits resulted in transfer of teacher-specific preferences or biases~\cite{cloud2026language}.
Importantly, this transfer was possible even after any explicit reference to the teacher's preference had been removed from the data. 
This highlights potential challenges for knowledge distillation---training smaller, more efficient student networks with data generated by a typically larger teacher network~\cite{hinton2015distilling}---because the same mechanism could transmit latent misalignment, hidden biases, or deliberately introduced backdoors \cite{betley2025emergent, hubinger2024sleeper, mohammadshahi2025distillation, souly2025poisoning}. 
Understanding subliminal learning is therefore essential to identify bounds when implicit transfer is possible and to prevent it when necessary.
Recent efforts in understanding subliminal learning have focused on LLMs~\cite{cloud2026language, kitkana2026sustained, aden2026subliminal} because of its relevance in application; 
however, we notice that this poses a central challenge in isolating the underlying mechanism, as a single vocabulary head produces both task-related and task-unrelated outputs.

Here, we address this shortcoming by studying a Multilayer Perceptron (MLP) learning MNIST as a controlled setup.
In the MLP, we can explicitly separate task-related and task-unrelated outputs into two heads: an auxiliary head (aux head) for task-unrelated logits and a classification head (class head) for task logits.
This preserves the central feature of subliminal learning---transfer through teacher outputs---while making the output structure directly controllable. 
We can thus test which output components enable subliminal transfer and when this transfer breaks down:
\begin{itemize}
    \item We show that, in principle, teacher and student do not need to share the same---or even similar---initializations in their hidden layers for subliminal learning to occur.
    \item We further show that this independence extends to network architecture: teacher and student architectures may differ, provided they satisfy loose conditions on comparable expressiveness.
    \item We identify compatibility of output layers (aux head / class head) and their stability during training as the main factor determining the success of subliminal learning and give a theoretical justification of this finding.
    \item We quantify perturbation resilience of output layers and failure modes in high dimensions supported with analytical theory.
\end{itemize}

\section{Related Work}

\citet{cloud2026language} introduced subliminal learning across two setups. 
In the LLM setting, a teacher with a target trait generates number sequences, code, or chain-of-thought traces on unrelated prompts; students fine-tuned on these outputs acquire the trait even after filtering. 
As a proof of concept, they replicate the effect in a small MLP on MNIST; the student matches the teacher's aux logits on random noise and still recovers digit classification, despite never seeing digits. 
They provide a theorem, which states: under shared initialization between teacher and student, a single step of the student's gradient descent on any teacher-generated output results in a parameter update that has a non-negative inner product with the teacher's gradient, except on a measure-zero orthogonal case; in other words, every step moves the student closer towards the teacher in parameter space. 
Their work was limited to strictly shared initialization, and cross-architecture transfer only works by first distilling the student on the teacher (\enquote{behavioral cloning}).

Follow-up work has pushed the limits by extending to multi-step training, showing that the alignment between trait and distillation gradients remains positive throughout training \cite{kitkana2026sustained}.
Subsequent work narrows the mechanism further, finding that around five percent of the tokens (divergence tokens) carry most of the signal \cite{schrodi2025towards}, challenging an earlier hypothesis that token entanglement and statistical logit leakage drive transmission \cite{zur2025token}. 
Another extension considers replacing the system's prompt with a steering vector on internal activations, thereby transmitting more complex biases and localizing the layers through which it propagates \cite{morgulis2026subliminal}. 

A related line of work is data-free knowledge distillation \cite{lopes2017datafree, yin2020dreaming}, where synthetic inputs are optimized to mimic the teacher's training distribution. Subliminal learning differs: the inputs are arbitrary noise, yet task-relevant information still transfers. Of all this work, only \citet{cloud2026language} and \citet{kitkana2026sustained} study the MLP setup, and treat it as a proof of concept rather than as an object of study.
 
\section{Conditions for Subliminal Learning}

\begin{figure}[ht]
    \centering
    \includegraphics[width=1\textwidth]{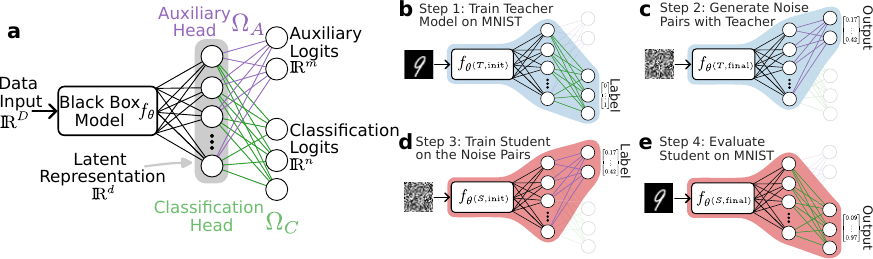}
    \caption{\textbf{Subliminal learning transfers task information through task-unrelated noise via compatible teacher--student output heads.}
    \textbf{(a)} We study models whose latent representation is connected to an aux head $\Omega_A$ producing task-unrelated auxiliary logits and a class head $\Omega_C$ producing task logits.
    \textbf{(b)} A teacher model is first trained on labeled MNIST~\cite{lecun2002gradient} using the class head.
    \textbf{(c)} The trained teacher is then queried on random-noise inputs, producing auxiliary outputs that form noise--target training pairs.
    \textbf{(d)} A student model is trained only on these noise pairs through its aux head.
    \textbf{(e)} Subliminal learning is observed when the resulting student, evaluated through its class head on MNIST, performs above chance despite never seeing MNIST labels during training.}
    \label{fig:subliminal_overview}
\end{figure}

\textbfplus{Subliminal learning setting}
We study subliminal learning in a controlled setting for small models on a classification task (\cref{fig:subliminal_overview}).

We consider a black box neural network model $f_\theta$ that maps input data $x^{(i)} \in \mathbb R^{D}$ into a latent-space $\mathbb R^d$ (\cref{fig:subliminal_overview}a). For convenience we shall call these latent representations $z^{(i)} = f_\theta(x^{(i)})$. The network possesses a split output head: First, the class head $\Omega_{C} \in \mathbb R^{n \times d}$ with a bias vector $b_C \in \mathbb R^n$ that maps the latent representation to logits $\Omega_C z^{(i)}+b_C$, converted into class-probabilities through $\mathrm{softmax}$. Second, the aux head $\Omega_A \in \mathbb R^{m \times d}$ acts as a linear readout $\Omega_A \cdot z^{(i)} + b_A$ of the latent-space onto $m$ auxiliary neurons. During subliminal learning we consider two different versions of this network: the teacher (T) and the student (S), distinguished by raised indices like $\theta^{(T,\text{init})}$ to refer to their weights, and to indicate their initial and final states before and after training.

The process of subliminal learning is the following: First, the teacher is trained on a classification task in a supervised manner (\cref{fig:subliminal_overview}b). During training only the output of the class head is used and the teacher's weights and class head are updated. Notably, the auxiliary neurons play no role in this part of the training. Hence the aux-head weights of the teacher remain in their initial state. In this work, our baseline setting uses an MLP with two hidden layers for both teacher and student and a latent dimension of $d=256$ (see~\cref{sec:methods}). Second, the student is trained, updating their weights $\theta^{(S)}$ and aux head $\Omega^{(S)}$, to match the teacher's readouts on the auxiliary neurons, given random noise data as input (\cref{fig:subliminal_overview}c,d). Notably, here the student class head is not involved in the training process and hence remains unchanged.

Subliminal learning is achieved, when the student has learned the original classification task of the teacher, without ever being trained on it. Specifically, evaluating the trained student on test data to predict class probabilities has to yield a (significantly) better than random accuracy (\cref{fig:subliminal_overview}e).

\textbfplus{When subliminal learning works}
We study subliminal learning from a theoretical perspective in \cref{sec:mb}, and derive the following general conditions for subliminal learning to work:
\begin{enumerate}
    \item The student network needs to learn and generalize the latent-representation of the teacher sufficiently well. We show the main requirement for this condition is compatibility of teacher and student aux heads.
    \item The class head of the teacher after training and the student class head need to be compatible.
\end{enumerate}
For the case of identical initialization of teacher and student, prior work by \citet{cloud2026language} has shown that the student weights will change in a similar direction as the teacher's during training, giving an explanation how the student can learn the teacher's latent-representation. Intuitively, the aux head acts as a random projection of the latent-space and thereby a student trying to match the teacher's outputs will (on average), update its learned representation in the correct direction. Here we show that the condition on identical initializations can be relaxed: If teacher and student share their initialization weights, but the aux head of the student is perturbed, the student can still learn a meaningful (though distorted) representation. Surprisingly, even a completely random aux head of the student can yield better than random performance, as it can recover during training (see \cref{sec:representation} for details). Even more surprisingly, when initialized with the same aux head but randomized hidden layers (even different architecture between teacher and student), representation learning can also work without any prior alignment, only by forcing the student to match the teacher's random aux-head projections, as long as the student's architecture is similarly expressive: it needs to be able to mimic the teacher's representation without overfitting on the noise-samples (see \cref{sec:mb_cross_arch}).

The second condition on the class heads derives from the fact that even a perfectly learned latent-representation will not yield meaningful classification performance, if the student class head is incompatible with the representation. Since the student class head is never trained, its initialization must be similar to the teacher after training $\Omega_C^{(S,\text{init})} \approx \Omega_C^{(T,\text{final})}$. The best performance is achieved if both match, and performance decreases quadratically with perturbation strength\footnote{Hence, it is somewhat robust to small perturbations}, but completely vanishes for random initialization (see \cref{sec:class-head}). 

\textbfplus{When subliminal learning fails}
The preceding theory turns subliminal learning into a set of testable predictions. The findings above suggest that head compatibility (both aux and class head) between teacher and student is the most pressing condition. If head compatibility is indeed the relevant condition, then shared hidden-layer initialization should not be necessary, class-head incompatibility should eliminate the effect, and aux-head mismatch should be recoverable only in favorable regimes with shared initialization.

Another practical failure mode arises from head drift during training: If the teacher's class head or the student's aux head changes significantly during training, performance will degrade. While one can show stability of both during training (see \cref{sec:aux_head_stability} and \cref{sec:mb_class_head_stability}) under typical conditions, gradient noise becomes relevant in high latent dimensions $d$, predicting class-head drift and failure of subliminal learning for very high dimensional representations (see \cref{sec:high_d_failure}). Finally, even with compatible heads, subliminal learning can fail if the student's architecture cannot learn (and generalize) the teacher's representation, due to underfitting or overfitting, putting a constraint on the similarity of expressiveness of teacher and student architectures for cross-architecture subliminal learning.

We now test these predictions empirically in the controlled split-head MNIST setting and across architectural, dataset, perturbation, and dimensionality variations.

\section{Results}

\begin{figure}[h]
    \centering
    \includegraphics[width=1\linewidth]{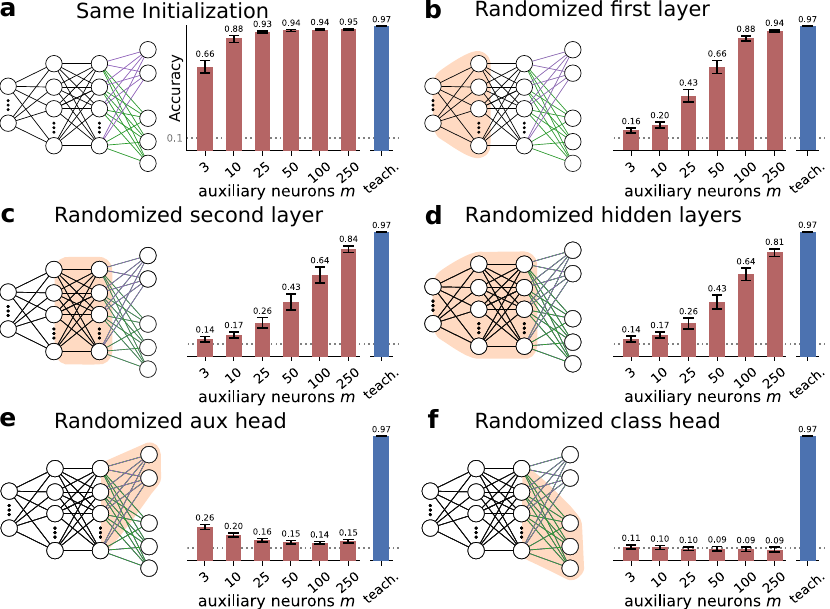}
    \caption{\textbf{Subliminal learning is robust to hidden-layer random initialization but fragile to output-head incompatibility.} 
    We test which shared student--teacher components are required for subliminal learning by randomly reinitializing different parts of the student before training. Orange regions indicate reinitialized components, red bars show student accuracy after training, and the blue bar shows teacher accuracy. Values show the mean over 20 seeds while error bars indicate bootstrapped 95\% confidence intervals.
    \textbf{(a)} With shared initialization, subliminal learning improves with the number of auxiliary neurons $m$ and approaches teacher performance. 
    \textbf{(b--d)} Reinitializing the first layer, second layer, or both hidden layers reduces efficiency, but subliminal learning re-emerges for sufficiently large $m$, showing that shared hidden-layer initialization is not necessary. 
    \textbf{(e)}~Reinitializing the aux head severely disrupts subliminal learning, indicating that the aux head must transmit a recoverable teacher signal. 
    \textbf{(f)} Reinitializing the class head eliminates subliminal learning across all tested $m$, showing that a compatible class readout is necessary for the recovered representation to yield task performance.}
    \label{fig:random_inits_barplots}
\end{figure}

\textbfplus{Layerwise dependency on shared initialization} We first ask which parts of a shared teacher--student initialization are actually required for subliminal learning. If shared initialization were required globally, reinitializing any component of the student should destroy the baseline effect (\cref{fig:random_inits_barplots}a). However, we observe that reinitializing the first layer, second layer, or both hidden layers reduces efficiency but does not eliminate subliminal learning when sufficiently many auxiliary neurons are available (see~\cref{fig:random_inits_barplots}b-d).

By contrast, reinitializing the aux head strongly suppresses the effect in our setting (see~\cref{fig:random_inits_barplots}e). This observation is consistent with the aux head acting as the readout channel that must remain similar between the models in order to enable transmission of a recoverable teacher signal. Our theory also predicts that combining hidden-layer randomization with a randomized aux head fully destroys recoverability (see \cref{sec:aux_head_stability}). Lastly, reinitializing the class head keeps performance at chance across all tested auxiliary dimensions (see~\cref{fig:random_inits_barplots}f, proven in \cref{sec:random_class_head_impossible}), showing that a compatible class readout is necessary for subliminal learning.

\begin{figure}[h]
    \centering
    \includegraphics[width=0.85\textwidth]{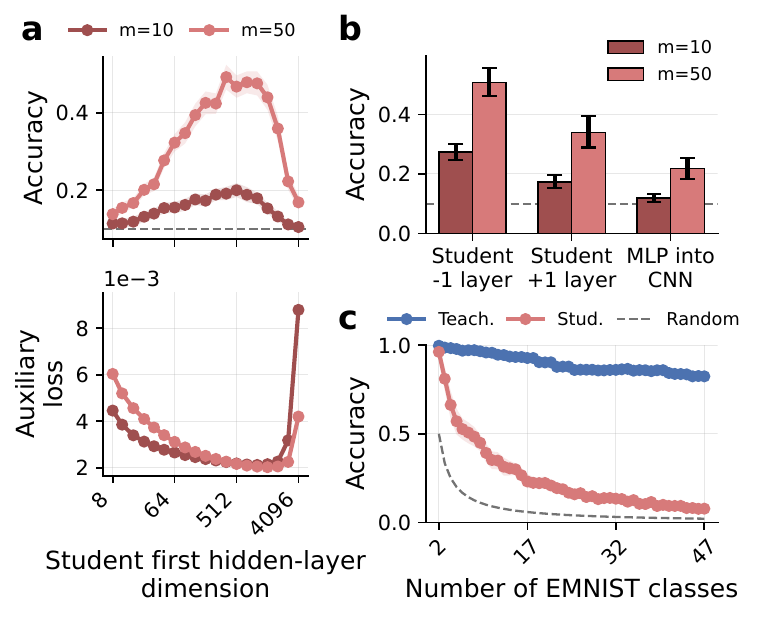}
    \caption{\textbf{Compatible output heads are sufficient for subliminal learning across architectural and dataset changes, but student capacity and task complexity determine recoverability.}
    We test subliminal learning when teacher and student share only the aux head and class heads while keeping a fixed teacher setup whereas the student hidden architecture or task is varied.
    \textbf{(a)} Varying the student first hidden-layer dimension reveals a nonmonotonic capacity dependence: underexpressive students achieve low accuracy, whereas overly wide students exhibit substantially higher auxiliary loss, indicating overfitting on noise batches, thus mostly fail to generalize the teacher signal.
    \textbf{(b)} Subliminal learning persists when the student MLP has one fewer or one additional hidden layer ($d=256$), and also for an MLP-to-CNN student (two convolutional layers into fully connected with $d=256$ with comparable parameter count, details in \cref{tab:mlp_cnn_hyperparameters}), showing that exact hidden-architecture matching is not required.
    \textbf{(c)} On EMNIST subsets with increasing numbers of classes, student ($m=50$) accuracy decreases faster than teacher accuracy, indicating that harder multiclass tasks require greater recoverability through the fixed auxiliary channel. Values show mean accuracy over 20 random seeds; error bars indicate bootstrapped 95\% confidence intervals.}
    \label{fig:fl-dim-cross-arch-emnist}
\end{figure}

\textbfplus{Architecture and dataset variations}
Next, we test whether subliminal learning requires matched student--teacher architectures or only compatible output heads. In these experiments, teacher and student share only the aux and class heads. Hidden layers are not matched and the student's architecture is varied. First, varying the student first hidden-layer dimension reveals a nonmonotonic capacity dependence because both underexpressive and overly wide students fail to recover the teacher signal, while approximately equally expressive students manage to do so (see~\cref{fig:fl-dim-cross-arch-emnist}a). Second, subliminal learning persists when the student MLP has one fewer or one additional hidden layer, and also when the student is replaced by a CNN of comparable parameter count (see~\cref{fig:fl-dim-cross-arch-emnist}b). Thus, exact hidden-architecture matching is not necessary for subliminal learning.

Third, on EMNIST subsets~\cite{cohen2017emnist}, student accuracy decreases as the number of classes increases, even though teacher accuracy remains high (see~\cref{fig:fl-dim-cross-arch-emnist}c). Harder multiclass tasks therefore require a stronger recoverable signal through the fixed aux head.

\begin{figure}[h]
    \centering
    \includegraphics[width=0.85\textwidth]{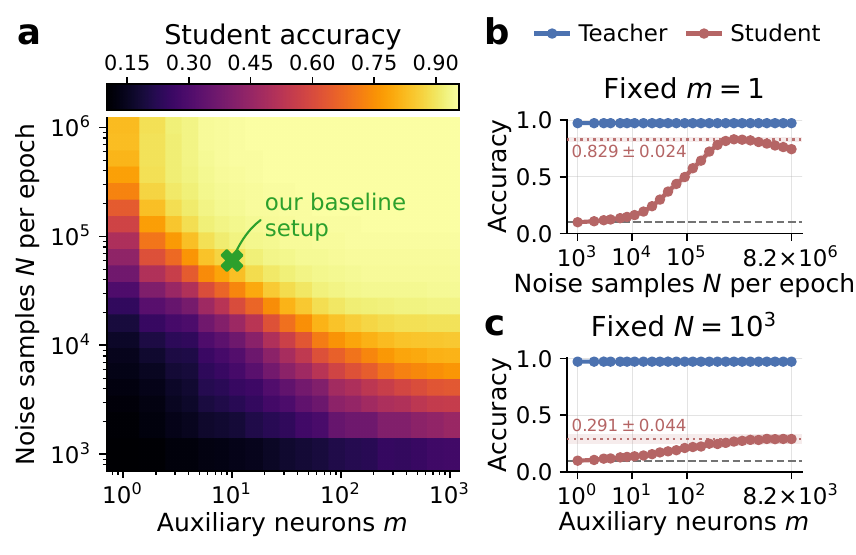}
    \caption{\textbf{Subliminal learning depends jointly on aux-head capacity and noise samples, with clear regimes of bottleneck, recovery, and saturation.}
    We vary the number of auxiliary neurons $m$ and the number of noise samples $N$ seen per student epoch to test how recoverability depends on aux-head capacity and noise exposure. 
    \textbf{(a)} Student accuracy across the $(m,N)$ plane. Increasing either $m$ or $N$ improves subliminal learning, but strong performance requires both sufficient aux-head capacity and sufficient noise samples.
    \textbf{(b)} With a fixed minimal aux head ($m=1$), increasing the number of noise samples initially improves student accuracy, but performance eventually peaks and then declines, indicating that additional noise cannot fully compensate for a severely capacity-limited aux head.
    \textbf{(c)} With a fixed noise budget ($N=10^3$), increasing the number of auxiliary neurons improves student accuracy only up to a plateau, showing that aux-head capacity alone cannot overcome limited noise samples.
    Values show mean accuracy over 20 random seeds; error bars indicate bootstrapped 95\% confidence intervals.}
    \label{fig:m_N_scaling}
\end{figure}

\newpage
\textbfplus{Scaling with auxiliary neurons and noise samples per epoch}
In general, when conditions for subliminal learning are met, information transfer depends on the strength of the auxiliary signal. The number of auxiliary neurons $m$ controls the number of random projections of the teacher latent representation available at each noise input, whereas the number of noise samples per epoch $N$ controls how many points of this projected function are observed. While $m$ increases the dimensionality of the signal per input, $N$ increases coverage of the projected noise distribution. We therefore expect subliminal learning to improve when either bottleneck is relaxed, but not without limit (compared to an optimal student that matches teacher accuracy within bootstrapped uncertainty, as described in \cref{tab:optimal_conditions}). High student accuracy requires both sufficient auxiliary projections and sufficient noise coverage (see~\cref{fig:m_N_scaling}a). Interestingly, though, even a single auxiliary projection can support strong subliminal learning when conducted on many noise inputs, but performance eventually peaks and declines at approximately $83\%$ accuracy (see~\cref{fig:m_N_scaling}b). With a fixed amount of noise samples per epoch, additional auxiliary neurons improve accuracy only up to a plateau at approximately $29\%$ accuracy, indicating that further auxiliary dimensions do not overcome the bottleneck of a noisy gradient signal (see~\cref{fig:m_N_scaling}c). Thus, $m$ and $N$ jointly control recoverability, but neither provides an unbounded route to subliminal learning.

\begin{figure}[ht]
    \centering
    \includegraphics[width=1\linewidth]{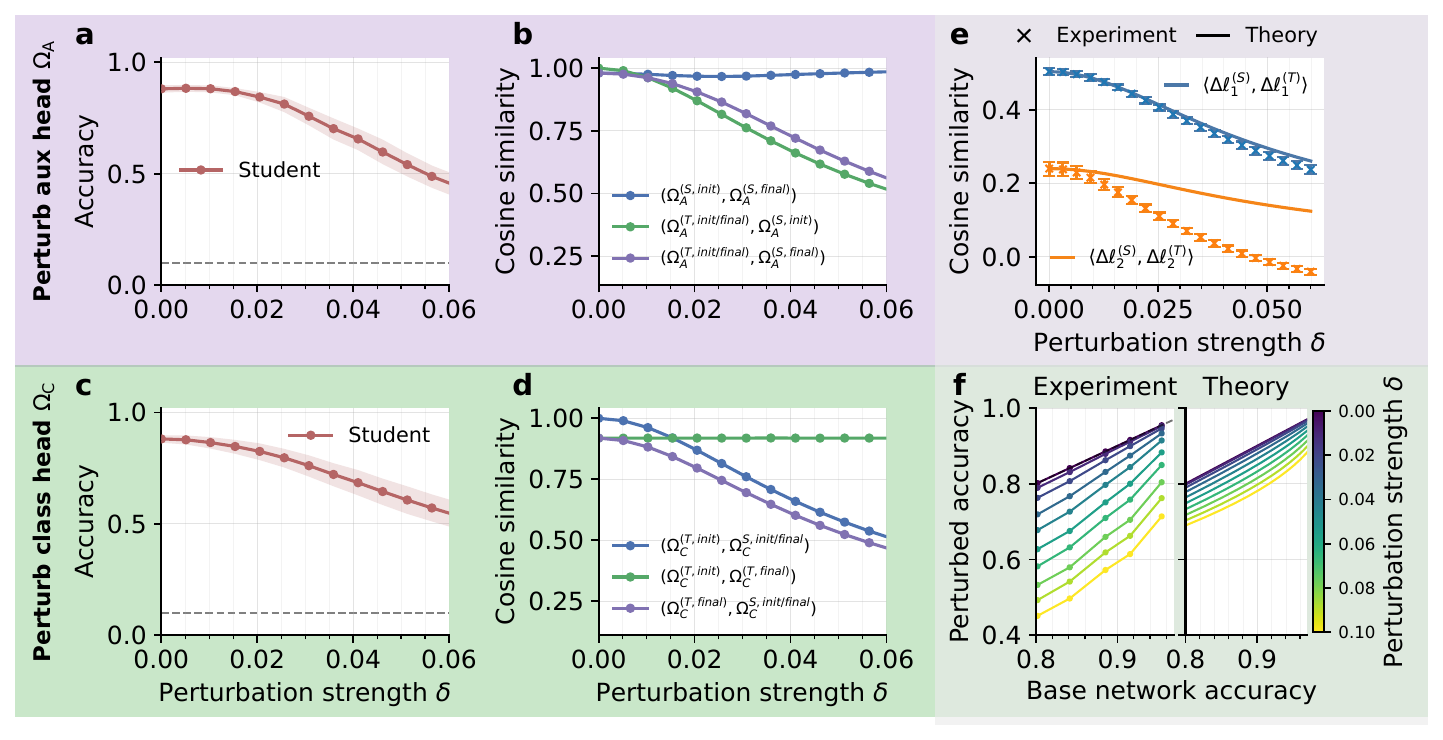}
    \caption{\textbf{Output-head perturbations reveal compatibility limits for subliminal learning and validate theory-derived robustness bounds.}
    Gaussian noise of strength $\delta$ is added to either the student's class or aux head before auxiliary training.
    \textbf{(a--d)} Perturbing either head reduces student accuracy and teacher--student head similarity, indicating that both readout and aux-head compatibility are required for successful signal recovery.
    Training partially restores aux-head, but not class-head, compatibility.
    \textbf{(e,f)} Post-hoc perturbations and hidden-update similarities are qualitatively captured by our theory: aux-head perturbations reduce hidden-update alignment, constrained by a theoretical upper bound (see \cref{eq:alignement_prediction}), while higher base accuracy increases class-head robustness, following an approximate scaling law (see \cref{eq:accuracy_scaling}).
    Means are over 20 seeds; shaded regions/error bars show bootstrapped 95\% CIs.}
    \label{fig:student_head_perturbation_sweep}
\end{figure}

\textbfplus{Perturbation analysis}
We next test whether head compatibility controls subliminal learning causally. Starting from our baseline setting, we perturb either the student aux head or the student class head before auxiliary training and vary the perturbation strength $\delta$. An aux-head perturbation affects the first readout condition. The student aux head is trained during auxiliary learning, so moderate perturbations can be partially corrected. Nevertheless, perturbing this head still reduces student accuracy and decreases teacher--student aux-head similarity (see~\cref{fig:student_head_perturbation_sweep}a,b). The final student aux head remains closer to the teacher aux head than the perturbed initialization, showing that auxiliary compatibility is recoverable in favorable regimes but not arbitrary, as supported by our theory in (\cref{sec:aux_head_stability}). 

A class-head perturbation changes the fixed readout used at evaluation, but this head receives no gradient during auxiliary training. Its effect is therefore direct and irrecoverable. Increasing $\delta$ steadily reduces student accuracy and decreases similarity between the student and teacher class heads (see ~\cref{fig:student_head_perturbation_sweep}c,d). This supports the second readout condition: even if the student recovers a useful representation, subliminal learning fails when the class head cannot decode it. This distinction explains why  aux-head and class-head perturbations are not equivalent: aux-head mismatch rotates the training signal that drives representation recovery, whereas class-head mismatch corrupts the final readout.

The response to perturbation can be explained by our theory. For aux-head perturbations, our theory (see \cref{sec:TheoryPerturbation}) predicts that perturbations reduce the similarity between teacher and student hidden-layer updates and force the student to learn a rotated teacher representation, and the empirical curves follow the same decay (see~\cref{fig:student_head_perturbation_sweep}e). Post-hoc perturbations of trained class heads show that higher-accuracy networks tolerate larger perturbations before accuracy collapses (see~\cref{fig:student_head_perturbation_sweep}f), consistent with the nonlinear relation between logits and classification accuracy (see \cref{sec:TheoryClassPerturb}). Additional sweeps at fixed aux-head perturbation (when the aux head is not trainable) show that increasing $N$ or $m$ can partly compensate for a skewed transfer, but the compensation is limited and nonmonotonic (see~\cref{fig:aux-p01-sweep}). We thus find that the compatibility between student and teacher heads becomes a necessary condition for subliminal learning.

\begin{figure}[h!]
    \centering
    \includegraphics[width=1\textwidth]{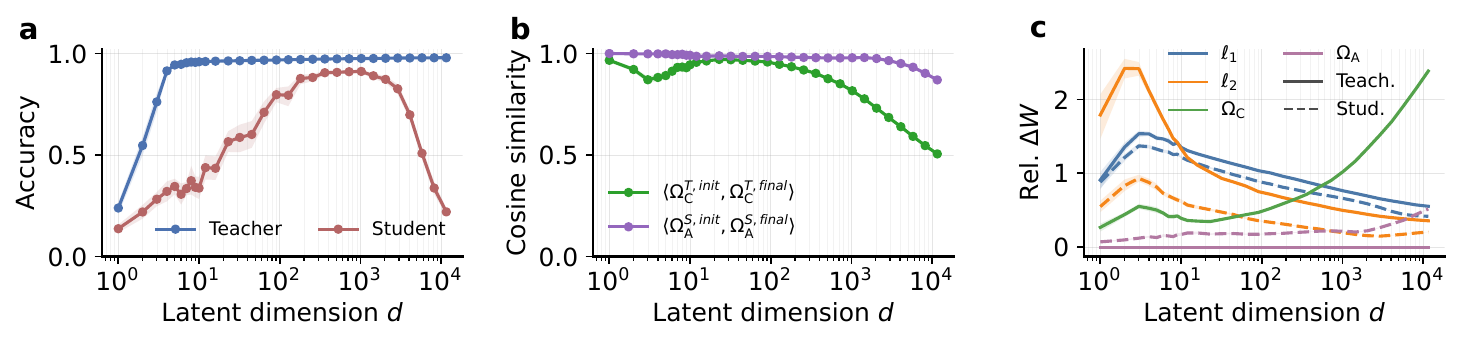}
    \caption{\textbf{Shared initialization alone does not guarantee subliminal learning, instead excessive latent dimensionality drives head drift and eventually breaks the effect.}
    We vary the shared latent dimension $d$ of teacher and student while keeping architecture and initialization otherwise identical.
    \textbf{(a)} Teacher accuracy quickly saturates with increasing $d$, whereas student accuracy first improves and then collapses at large $d$, showing that subliminal learning can fail even under fully shared initialization.
    \textbf{(b)} This failure coincides with increasing head drift: as $d$ grows, the teacher class head moves progressively farther from its initialization, while the student aux head remains comparatively stable over a broad range before also drifting at large $d$.
    \textbf{(c)} Relative weight changes across layers show that large latent dimensions disproportionately amplify head updates, in particular for the teacher class and student aux head, supporting the view that loss of head compatibility drives the breakdown of subliminal learning.
    Values show means over 20 random seeds; shaded regions indicate bootstrapped 95\% confidence intervals.}
    \label{fig:d_sweep_summary}
\end{figure}

\textbfplus{Representation dimensionality as a failure mode}
Finally, we test whether shared initialization is sufficient once architecture, heads, and all hidden layers are matched. We vary the shared latent dimension $d$ while keeping teacher and student identical. If shared initialization were sufficient, increasing $d$ should not destroy subliminal learning. Instead, we observe a nonmonotonic dependence on $d$ (see~\cref{fig:d_sweep_summary}a). Teacher accuracy quickly saturates, while student accuracy first improves and then collapses for large latent dimensions. Thus, even fully shared initialization can fail.

The collapse in accuracy coincides with head drift. As $d$ increases, the teacher class head moves farther from its initialization, while the student aux head also starts to drift at large $d$ (see~\cref{fig:d_sweep_summary}b,c and \cref{sec:high_d_failure} for the theoretical reasoning). Both effects reduce head compatibility, but they do not contribute equally. To separate them, we repeat the dimensionality sweep with fixed aux head, fixed class head, and both. When the aux head is fixed, the teacher class head still drifts, but the high-$d$ accuracy drop is much slower. When the class head is fixed, the drop remains severe once the aux head drifts. This identifies aux-head drift as the dominant failure mode in the high-dimensional regime. Class-head drift accounts for a smaller degradation, as seen by comparing the aux-head-fixed condition to the both-heads-fixed condition. Even when both heads are fixed, a residual dimensionality-dependent drop remains, showing that latent dimensionality can also impair subliminal learning through the hidden representation itself. Shared initialization therefore provides only initial compatibility; training with a (very) high-dimensional latent space can still destroy recoverability, primarily through random drift of the aux head due to a decaying signal-to-noise ratio of its training gradients (see~\cref{fig:d_sweep_ablations}, theory in \cref{sec:high_d_failure}).

\section{Methods}
\label{sec:methods}
\textbfplus{Baseline setup}
We study subliminal learning in a split-head teacher--student setting. Unless stated otherwise, teacher and student are initialized identically before teacher training. Both models are two-hidden-layer MLPs with hidden width $d_{\ell_1}=d_{\ell_2}=256$, ReLU activations, a class head $\Omega_C$ with output dimension equal to the number of classes, and an aux head $\Omega_A$ with dimension $m=10$. All linear layers use PyTorch's default initialization \cite{paszke2019pytorch}. For a layer with $d_{\mathrm{in}}$ input connections per output unit, weights and biases are sampled from $\mathcal{U}(-d_{\mathrm{in}}^{-1/2}, d_{\mathrm{in}}^{-1/2})$. Full initialization, optimizer, and preprocessing details are reported in~\cref{tab:baseline_hyperparameters}.

The teacher is trained on labeled data using cross-entropy loss on the classification logits. The aux-head weights are not directly optimized during teacher training, but the auxiliary outputs change through the trained representation. The student remains at the shared initialization until auxiliary training. During auxiliary training, the student is trained only on synthetic noise inputs by matching the teacher auxiliary logits with mean-squared error~\cite{kim2021comparing}. The student class head receives no gradient during auxiliary training. We evaluate subliminal learning by the final student classification accuracy on the held-out test set.

Unless stated otherwise, experiments use the MNIST dataset, uniform noise $\mathcal{U}(-1,1)$, noise batch size $B_{\mathrm{noise}}=1000$, $S_{\mathrm{noise}}=60$ gradient steps per student epoch, and five teacher and student epochs. The number of noise samples per student epoch is $N = B_{\mathrm{noise}} S_{\mathrm{noise}}$. Each batch consists of independently sampled noise inputs. Main-text results are averaged over 20 random seeds; confidence intervals are bootstrapped over seeds.

\textbfplus{Experimental degrees of freedom}
We vary the baseline setup along four axes. First, teacher--student initialization can be shared or randomized independently per layer, which isolates hidden-layer, aux-head, and class-head compatibility. Second, layers and heads can be frozen or trained independently, which enables fixed-head ablations. Third, we sweep the aux-head capacity $m$ and the number of noise samples $N$ to measure how both quantities control recoverability. Fourth, we vary architecture and task complexity, including MLPs with different depth or width, MLP--CNN experiments, and EMNIST subsets with different numbers of classes. Experiment-specific deviations from the baseline are reported in~\cref{apx:experiment_configs}. For CNN students, we use spatially correlated Perlin noise~\cite{perlin1985image} with resolution 8, because independent pixel noise does not reliably excite convolutional features (see~\cref{fig:perlin_noise}).
For perturbation experiments, selected weights and biases are perturbed by additive Gaussian noise with scale $\delta$ before auxiliary training. 

\section{Discussion}
In our work, we use a controlled MNIST setting to study the fundamental mechanism underlying subliminal learning in an MLP. We set out to determine what is actually required for subliminal learning here, and our central finding reframes the phenomenon: subliminal learning is governed by output-head compatibility between teacher and student, not by shared hidden layer initialization as previously argued~\cite{cloud2026language}. A sufficiently expressive student---across reinitialization of hidden layers, modified depth and even MLP-to-CNN transfer---can recover the teacher's latent representation through a compatible aux head. If a compatible class head can read that representation out, subliminal learning works.

The two heads fulfill two distinct roles. The student's aux head plays a crucial role in recovering the teacher's latent representation: it provides random projections of the latent space, so any geometric mismatch between student and teacher aux head rotates the student's training signal, explored in detail in \cref{sec:TheoryPerturbation}, causing a substantial drop in knowledge transfer that the student can only partially recover during training. The teacher's class head, in turn, must remain stable during training, a condition which we examine in detail in \cref{sec:class-head}. When either head fails to meet its respective condition, subliminal learning breaks down in predictable ways.

The different roles of the heads explain much of our empirical results. The student's aux head is updated during student training and can partially correct an initial mismatch---the final student aux head consistently moves closer to the teacher's than to the perturbed initialization (\cref{fig:student_head_perturbation_sweep})---provided the rest of the network supplies a useful gradient signal (see \cref{sec:aux_head_stability}). The class head, in contrast, receives no gradient during student training, so any incompatibility---from independent initialization, perturbation, or teacher-side drift---persists irrecoverably. This asymmetry predicts the patterns we observe: hidden-layer randomization is tolerable because the student can recover a representation the shared heads can encode and decode; class-head perturbations are better tolerated by higher-accuracy networks because well-separated logits do not flip the argmax under small rotations; and aux-head drift dominates the high-dimensional failure mode (\cref{fig:d_sweep_summary}, \cref{fig:d_sweep_ablations}), because the class head only inherits teacher-side drift while a perturbed student aux head actively rotates the student's learned representation away from the teacher's during training.

Our framework predicts under which conditions subliminal learning breaks. 
Cross-architecture transfer succeeds only when the student has a similar expressiveness as the teacher. 
If it is too narrow, it cannot match the teacher's projections; if it is too wide, it overfits the noise (\cref{fig:fl-dim-cross-arch-emnist}a).
In a task with more classes, student accuracy degrades faster than the teacher accuracy as they demand greater recoverability through a fixed auxiliary channel (\cref{fig:fl-dim-cross-arch-emnist}c).
The number of auxiliary neurons and the number of noise samples used for training are jointly necessary but are mutually irreplaceable: a single auxiliary projection can support strong transfer only when paired with sufficient noise samples, and a high number of auxiliary neurons with enough noise samples (\cref{fig:m_N_scaling}). 
Finally, high latent dimensionality can break subliminal learning even under fully shared initialization, because gradient noise drives the teacher's class head and the student's aux head away from compatibility (\cref{fig:d_sweep_summary}).

\textbfplus{Limitations and future work}
Some observations remain incompletely explained by our current theory. 
Specifically, we find a dip in accuracy upon increasing the latent dimension at $d=m$ (\cref{fig:d_sweep_summary})---possibly connected to the aux-head size (\cref{fig:d_sweep_emnist})---that we suspect is linked to a breakdown of our assumption of orthogonal projections. 
Similarly, our theory does not predict the initial reduction in accuracy when increasing $m$ under a random initialized aux head (\cref{fig:random_inits_barplots}e), though we can at least explain why it does not increase (\cref{sec:TheoryPerturbation}). 
More importantly, our work considered subliminal learning in a controlled MNIST setting with an explicit aux head compared to large-language models, where task-relevant and task-unrelated information flow through the same vocabulary head~\cite{cloud2026language}. 
We expect, however, that our framework can be generalized to LLMs in the future, because the vocabulary layer projects distinct concepts onto largely non-overlapping token regions \cite{geva2022transformer, elhage2022toy}, allowing trait-relevant and distillation-relevant tokens to act as approximately independent linear readouts of the model's latent space. Under this mapping, the set of distillation-relevant tokens plays the role of the aux head and the set of trait-relevant tokens plays the role of the class head. 
Because this disentanglement is only implicit, the compatibility constraint may be stronger than in our controlled setting, consistent with the empirical conditions reported for LLM subliminal learning.

While the direct risks of subliminal learning in our controlled MLP setup are limited, the general phenomenon poses a broader concern: distillation can transfer biases or hidden traits under substantially weaker conditions than previously thought. Prior work tied subliminal learning to closely matched teacher--student initialization, which bounds the risk to distillation between near-identical models. Our results remove this bound: transfer persists under randomized hidden layers, different student depth, and even across architecture families. The operative condition is compatibility at the output interface, not similarity of internal representations.
A natural and important next step is to test whether this loosened condition also extends to LLMs, where the vocabulary layer would act as the output interface. 
\citet{lee2025shared} show that models within the same family often share similar relative token orientations, despite differing embedding dimensions; in untied Llama-3 models, they find high correlations specifically in the unembedding space. In this regime, our finding predicts that hidden-layer diversity no longer provides a barrier to trait inheritance, and pipelines distilling a shared teacher into a heterogeneous family of students may inherit the same risk profile. Across model families with different tokenizers, unembedding matrices differ structurally and have no canonical alignment~\cite{goddard2025training}; whether subliminal transfer still occurs there remains open.

We conclude that subliminal learning is not a mysterious artifact of shared initialization but a controlled consequence of output-head compatibility and of reshaping latent projections using noise-based training---a reframing that makes implicit information transfer in larger teacher-student systems mechanistically tractable.

\begin{ack}

We are grateful to Johannes Zierenberg for his extensive feedback on this manuscript. We further thank Lisa Beinborn and Alexander Ecker for their valuable input and the fruitful discussions on this topic. Our thanks also go to the rest of the Priesemann Group for their thoughtful comments and feedback on this work. Finally, we thank the Danni Heineman Stiftung for co-funding the workshop Perspectives in Complex Self-Organized Systems, where this project was initiated.
V.P. was funded via the MBExC by the Deutsche Forschungsgemeinschaft (DFG, German Research Foundation) under Germany’s Excellence Strategy-EXC 2067/1-390729940. V.N. and V.P. were supported and funded by the DFG – GRK2906 – project number 502807174. V.P. received funding from the DFG via the SFB 1528 “Cognition of Interaction” - project-ID 454648639.
\end{ack}

\bibliography{bibliography}

\clearpage
\appendix

\section{Mathematical Background}

\label{sec:mb}

\subsection{Subliminal Learning Setting}
\label{sec:mb_settig}
We consider a "black box" neural network model $f_\theta$ that maps an input vector $x^{(i)} \in \mathbb R^{D}$ into a latent space $\mathbb R^d$. For convenience we shall call these latent representations $z^{(i)} = f_\theta(x^{(i)})$. We then have a class head $\Omega_{C} \in \mathbb R^{n \times d}$ that maps the latent representation onto logits for the $n$ classes $c=1..n$ and then a probability vector $p(c|z^{(i)}) \in \mathbb R^n$  via softmax. At the same time we also have an aux head $\Omega_A \in \mathbb R^{m \times d}$ that acts as a linear readout of the latent-space on $m$ auxiliary neurons. The resulting read-outs of the latent-space are
\begin{align}
 \text{class-head output:}\;\;\; &\begin{cases} &p(c|z^{(i)}) = \mathrm{softmax}\left(\Omega_{C} \cdot z^{(i)} + b_C\right) \text{}\end{cases}\\
 \text{aux-head output:}\;\;\; &\begin{cases} &\Omega_A \cdot z^{(i)} + b_A\end{cases}\;.
\end{align}
Here $b_A \in \mathbb R^m$ and $b_C \in \mathbb R^n$ are the biases for the auxiliary neurons and class head respectively. During subliminal learning we consider two different versions of this network: the teacher and the student. To distinguish the weights of both, we will use raised indices like $\theta^{(T,\text{init})}$ to refer to teacher (T) and student (S) weights, as well as their initial and final states before and after training.

The process of subliminal learning is as follows: First, the teacher is trained on a classification task in a supervised manner. Let $x^{(i)}$ ($i=1,..,N$) be the data with known class-labels $c_i$. During training only the output of the class head is used and the teacher's weights and class head are updated
\begin{equation}
    \left(\theta^{(T,\text{init})},\Omega_C^{(T,\text{init})},b_C^{(T,\text{init)}}\right) \rightarrow \left(\theta^{(T,\text{final})},\Omega_C^{(T,\text{final})},b_C^{(T,\text{final)}}\right)\;.
\end{equation}
Notably, the auxiliary neurons play no role in this part of the training. Hence the aux-head's weights and biases do not change at all and we have 
\begin{align}
\Omega_A^{(T,\text{init})} &= \Omega_A^{(T,\text{final})}, & b_A^{(T,\text{init)}}&=b_A^{(T,\text{final)}}\;.
\end{align}

In this work we use cross-entropy loss to train the network for this classification task, that is

$$\mathcal{L}=-\sum_{i=1}^{N}\log{\left(p^{(T)}\left(c=c_i\Big|f_{\theta^{(T)}}(x^{(i)}\right)\right)}\;,$$
though other losses (as long as they are convex) are possible. Second, the student is trained to match the teacher's outputs of the auxiliary neurons. These outputs are usually generated by noise data $n^{(i)}$. The aim is to match the linear readout from the auxiliary neurons of teacher and student 
\begin{equation}\Omega_A^{(S)} \cdot f_{\theta^{(S)}}(n^{(i)}) + b_A^{(S)} \approx \Omega_A^{(T)} \cdot f_{\theta^{(T)}}(n^{(i)}) + b_A^{(T)}\;.\end{equation}
In this work we use MSE loss to train the student, given it is the most natural to recover the linear structure of the latent-space, though other losses are possible as well. During training the student weights and aux head are updated 
\begin{equation}
    \left(\theta^{(S,\text{init})},\Omega_A^{(S,\text{init})},b_A^{(S,\text{init)}}\right) \rightarrow \left(\theta^{(S,\text{final})},\Omega_A^{(S,\text{final})},b_A^{(S,\text{final)}}\right)\;.
\end{equation}
Notably, this time the student class head is not involved in the training process, and hence remains unchanged: 
\begin{align}
    \Omega_C^{(S,\text{init})}&=\Omega_C^{(S,\text{final})}, & b_C^{(S,\text{init})}&=b_C^{(S,\text{final})}\;.
\end{align}
The subliminal learning effect is now achieved, when the student has "learned" the original classification task of the teacher, without ever being trained at it. Specifically, evaluating the student on data to predict class probabilities
\begin{equation}
    p^{(S)}(c|x)=\mathrm{softmax}\left(\Omega_{C}^{(S,\text{final})}\cdot f_{\theta^{(S,\text{final})}}(x) + b_C^{(S,\text{final})}\right)\;,
\end{equation}
yields a (significantly) better than random accuracy. In the following it will be discussed when subliminal learning can and cannot happen, depending on the initialization states and training procedures of both networks.

\subsection{Conditions for Subliminal Learning}
\label{sec:mb_conditions}
The following two conditions are sufficient for subliminal learning to work:
\begin{enumerate}
    \item The student network needs to learn the latent-representation of the teacher sufficiently well. It needs to generalize the prediction of the teacher latent output from noise inputs to the data samples $x$
    \begin{equation}
        f_{\theta^{(S,\text{final})}}(x)\approx f_{\theta^{(T,\text{final})}}(x)\;.
        \label{eq:representation_condition}
    \end{equation}
    \item The final class head of the teacher and the student class head need to be sufficiently close 
    \begin{equation}
        \Omega_C^{(T,\text{final})} \approx \Omega_{C}^{(S,\text{init})}\;.
        \label{eq:classheadcondition}
    \end{equation}
\end{enumerate}
If the student has learned the teacher's latent-output and the class head of both is similar enough, the student's classification probabilities will be close to the teacher's. 

Conversely, having an incorrect class head will degrade performance or outright erase it (see \cref{sec:class-head}), making \eqref{eq:classheadcondition} a necessary condition for subliminal learning as well. However, if the minimal aim for subliminal learning is only to achieve a "better than random" classification performance of the student, we can still relax the condition \eqref{eq:representation_condition} somewhat, giving us the necessary condition that teacher and student representation have to match on a sub-space $V \subset \mathbb R^d$ of the latent-space that does not lie in $\mathrm{ker}(\Omega_C^{(S,\text{init})})$
\begin{equation}
    P_Vf_{\theta^{(S,\text{final})}}(x)\approx P_Vf_{\theta^{(T,\text{final})}}(x)\;
\end{equation}
to at least recover some of the teacher's classification signal (see \cref{sec:representation}).

In order to properly assess under which conditions subliminal learning can work, it is important to quantify what exactly "$\approx$" means in the equations above. We will address this in the following sections.

\subsection{Latent Representation Condition}
\label{sec:representation}
We first focus on assessing when the former condition \eqref{eq:representation_condition} can be achieved.

\subsubsection{Representation learning from shared initialization}
In the case of a shared initialization of student and teacher weights $\theta^{(S,\text{init})}=\theta^{(T,\text{init})} =: \theta^{(0)}$ and aux head $\Omega_A^{(S,\text{init})}=\Omega_A^{(T,\text{init})}=:\Omega_A$ prior work by \citet{cloud2026language}, has established a "subliminal learning theorem" that guarantees teacher and student weights change in a similar direction at the begin of the student's training, explaining how the student learns to reproduce the teacher's internal representation from the auxiliary outputs. We shall briefly reproduce their arguments here in our setting to subsequently expand on them.

Let $\Delta\theta^{(T)} = \theta^{(T,\text{final})}-\theta^{(0)} $ be the change of the teacher's weights during training (for simplicity of notation we consider these weights to be a single, high dimensional parameter vector). Using a linearization argument, common in the literature studying neural tangent kernels \cite{linearization}, we want to predict how the student weights will change at the start of training, assuming both teacher and student do not change their weights too much. Let $n^{(i)}$ be a noise sample for the training of the student. We approximate\begin{align}f_{\theta^{(T,\text{final})}}(n^{(i)}) &\approx f_{\theta^{(0)}}(n^{(i)}) + J^{(i)}\Delta\theta^{(T)} & J^{(i)}&=\frac{\partial f_{\theta}(n^{(i)})}{\partial \theta}|_{\theta = \theta^{(0)}}\end{align}
with $J^{(i)}$ being the Jacobian of the network output w.r.t. all parameters, when evaluated at the noise sample. Similarly, we have for the student
\begin{align}f_{\theta^{(S,\text{final})}}(n^{(i)}) &\approx f_{\theta^{(0)}}(n^{(i)}) + J^{(i)}\Delta\theta^{(S)}\end{align}

For this singular noise sample, the MSE loss on the aux head reads
\begin{align}\mathcal L(\theta^{(S)}) &= \frac{1}{2}||(\Omega_A f_{\theta^{(S)}}(n^{(i)}) - b_A) - (\Omega_A f_{\theta^{(T)}}(n^{(i)}) - b_A)||_2^2\\
 &\approx \frac{1}{2}||\Omega_A J^{(i)}\cdot ((\theta^{(S)}-\theta^{(0)}) - \Delta \theta^{(T)})||_2^2.
\end{align}
The gradient at the beginning of training evaluates to
\begin{equation}
    -\nabla_{\theta^{(S)}} \mathcal L = (\Omega_A J^{(i)})^\intercal(\Omega_A J^{(i)}) \Delta \theta^{(T)}\;.
\end{equation}
When the student's weights are updated during training, we have
\begin{equation}
    \Delta \theta^{(S)} \propto -\nabla_{\theta^{(S)}} \mathcal{L}\;.
\end{equation}
Incorporating a batch over multiple samples $i=1,..,B$, this reads
\begin{equation}
    \Delta \theta^{(S)} \propto \frac{1}{B}\left(\sum_{i=1}^B (\Omega_A J^{(i)})^\intercal(\Omega_A J^{(i)})\right) \Delta \theta^{(T)}\;.
\end{equation}
Importantly $(\Omega_A J^{(i)})^\intercal(\Omega_A J^{(i)})$ is a positive semi-definite symmetric matrix (and so is the sum over a batch of noise samples), and therefore \begin{equation}
    \langle\Delta \theta^{(S)},\Delta \theta^{(T)}\rangle > 0
    \label{eq:subliminal_learning_thm}
\end{equation}
almost surely\footnote{For $\langle\Delta \theta^{(S)},\Delta \theta^{(T)}\rangle = 0$, the image of the gradient $J^{(i)}$ needs lay in the kernel of the aux head, which is almost surely not the case for random noise vectors as input.}, indicating that the student weights move in the direction of the teacher's final weights at the beginning of training, recovering the subliminal learning theorem from \citet{cloud2026language}. Since student weights move in the direction of the teacher's weights, the student's learned representation will become more similar to the teacher's during training, and the student will learn an approximation of the teacher's latent representation.

Note that the matrix connecting the teacher and student weights has the form $(J^{(i)})^\intercal (\Omega_A^\intercal \Omega_A) J^{(i)}$. The central matrix $\Omega_A^\intercal \Omega_A$ is data-independent and in the case of high latent-dimension $d \gg 1$ can be interpreted as an (approximate) orthogonal projection from the latent-space onto the span of the (row)-vectors of the aux-head matrix. To see this, write \begin{align}\Omega_A &=  \begin{bmatrix}
           \alpha^\intercal_{1} \\
           \vdots \\
           \alpha^\intercal_{m}
\end{bmatrix}\;,\end{align}
with vectors $\alpha_i \in \mathbb R^d$. In practice, for variance preserving initialization schemes~\cite{kaiming2015delving}, the components of the $\alpha_i$ vectors will often be initialized with centered random values of scale $\frac{1}{\sqrt{d}}$, giving the vectors an approximate length of $\mathbb E(||\alpha_i||_2^2) =:\beta \in \mathcal{O}(1)$, independent of $d$. Importantly, for high latent dimensions $d \gg 1$, these random vectors become pairwise (approximately) orthogonal since their cosine similarity scales as $\frac{1}{\sqrt{d}}$. Hence, for typical initializations and $m \ll d$, $\Omega_A^\intercal \Omega_A$ will effectively become a random orthogonal projection of the latent-space onto an $m$-dimensional sub-space (up to a the constant pre-factor $\beta$, that is depending on the initialization scheme\footnote{For example, for normal distributed initial weights of std $\frac{1}{\sqrt{d}}$ one has $\beta=1$; for uniform noise on $\pm \frac{1}{\sqrt{d}}$ it holds $\beta=1/3$} and we can effectively write 
\begin{equation}
    \Delta \theta^{(S)} \propto \frac{\beta}{B}\left(\sum_{i=1}^B  (J^{(i)})^\intercal PJ^{(i)}\right) \Delta \theta^{(T)}
\end{equation}
with a random orthogonal projection $P$ with rank $m$. Notably, the higher $m$, the more gradient signal towards the teacher's representation is recovered, though even for $m=1$, the gradient will point in the right direction, explaining why even in this extreme case subliminal-learning can work, provided that teacher and student share an initialization.

\subsubsection{Representation learning with a perturbed aux head}
\label{sec:TheoryPerturbation}

Next we consider a setting in which now the aux head of the student and teacher are not exactly the same, but rather the student's aux head is perturbed by noise 
\begin{equation}
    \Omega_A^{(S,\text{init})} = \Omega_A^{(T,\text{init})}+\epsilon\;.
\end{equation}
Such a training initialization scenario might be imaginable, if for example the teacher's final weights are not fully or only approximately known (but the teacher can be evaluated). We consider $\epsilon_{ij} \overset{\text{iid}}{\sim} N(0,\delta^2)$ with the aim to investigate the resilience of the student's training w.r.t the noise strength $\delta$. In the case $d \gg m$, we can again assume the row-vectors $\alpha_i$ of $\Omega_A$ to be approximately orthogonal. Additionally, decomposing the perturbation matrix $\epsilon$ into row-vectors 
\begin{align}\epsilon &=  \sqrt{d} \delta \begin{bmatrix}
           h^\intercal_{1} \\
           \vdots \\
           h^\intercal_{m}
\end{bmatrix}\;,\end{align}
each of the (rescaled) row-vectors has approximately\footnote{$||h_i||_2^2 \sim \frac{1}{d} \chi^2_d$} norm $1$ and is orthogonal to the other row vectors (as well as the $\alpha_i$). Retracing the steps of the previous derivation, but considering first order terms in the perturbation strength of the student aux head, we arrive at the following expression for the loss gradient $\nabla_{\theta^{(S)}}\mathcal{L}$, given a single noise sample $n^{(i)}$:

\begin{equation}
    - \nabla_{\theta^{(S)}} \mathcal L \propto \sqrt{\beta}(J^{(i)})^\intercal[\underbrace{\sqrt{\beta} P}_{\text{orig. term}} - \underbrace{( \sqrt{d} \delta)\cdot Q}_{\text{shear term}}] J^{(i)} \Delta \theta^{(T)} - \underbrace{(\sqrt{d} \delta) \cdot (J^{(i)})^\intercal Q^\intercal f_{\theta^{(0)}}(n^{(i)})}_{\text{gradient noise}},
    \label{eq:auxpertgrad}
\end{equation}
where the matrix $Q \in \mathbb R^{d \times d}$ is defined by 
\begin{equation}
    \frac{1}{\sqrt{d\beta} \delta}\epsilon^\intercal \Omega_A\;.
\end{equation}
$Q$ generates a shear-motion in the weight space. It projects a latent-space vector onto the (normalized) aux-head vectors $\alpha_i$ and maps the result along the direction of the orthogonal (normalized) noise-vectors $q_i$. Both the random projection $P$ and $Q$ have matrix-norm $\approx 1$ and act on the same sub-space of the latent-space, spanned by the aux-head vectors $\alpha_i$. Hence, the comparing factor 
\begin{equation}
r=\frac{\sqrt{d}\delta}{\sqrt{\beta}}    
\end{equation}
determines, how the teacher's signal updates the student's weights. If $r \ll 1$ the original subliminal learning signal dominates and teacher and student weight updates point in a similar direction. If $r \gg 1$, the shearing effect dominates and the student's weight updates are performed in random orthogonal directions, preventing the student from learning the teacher's representation. It should be noted, that other than the second noise term, this shear effect does not average out, when considering the averaged gradient over a batch of noise samples, and so we find the perturbed aux-head student weight updates to be
\begin{equation} 
\Delta \theta^{(S,\delta)} \propto \frac{\beta}{B}\left(\sum_{i=1}^B  (J^{(i)})^\intercal (P- r Q)J^{(i)}\right) \Delta \theta^{(T)}\;.
\end{equation}
in first order of $\delta$. Importantly, since $Q$ is not symmetric, for sufficiently large perturbations strength $\delta$, the previous subliminal learning theorem, no longer has to hold as it is possible that $\langle \Delta \theta^{(S)},\Delta \theta^{(T)}\rangle \leq 0$. Assuming small perturbation strength $\delta$, we can calculate the dependence of the average cosine-similarity of student and teacher updates for many runs to be
\begin{equation}
    \mathbb{E}(\mathrm{cos}(\Delta \theta^{(S,\delta)},\Delta \theta^{(T)})) = \frac{ \mathrm{cos}(\Delta \theta^{(S)},\Delta \theta^{(T)})}{\sqrt{1+r^2}}\;.
    \label{eq:alignement_prediction}
\end{equation}
This is assuming $\delta$ to be small and importantly, that the noisy gradient term in \eqref{eq:auxpertgrad}, can be neglected. It presents as an upper bound for the (average) similarity between teacher and student weight updates (at the begin of the student training), compared to the alignment of teacher and student weight updates when training with the un-perturbed aux head. We conclude that a student can still learn the correct representation of a perturbed aux head for small perturbations, as long as $r \ll 1$, but for higher perturbations the student will learn a systematically distorted representation. Therefore subliminal learning is only somewhat robust to a perturbed aux head. Importantly, in the regime $m \ll d$ the strength of the misalignment of, teacher and student, does not directly depend on the number of auxiliary neurons - suggesting that one cannot escape learning a systematically misaligned by simply adding more (slightly distorted) latent-space readouts.

\begin{figure}[ht]
    \centering
    \includegraphics[width=1.0\textwidth]{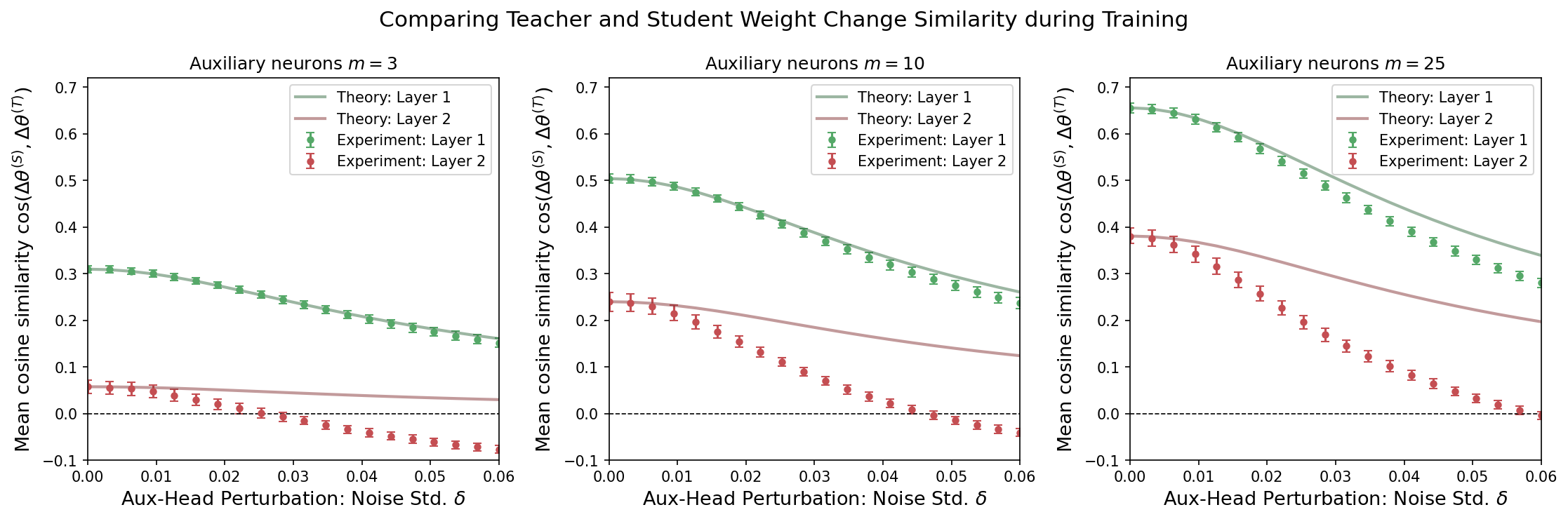}
    \caption{\textbf{Perturbing the student's aux head reduces the similarity between teacher and student hidden-layer updates forcing the student to learn a rotated teacher representation} Without perturbation we observe an alignment of teacher and student weight-changes, as presented in eq. \eqref{eq:subliminal_learning_thm}. Training was performed for a single epoch to capture changes at the beginning of training. The theory describing an upper bound for the decay in alignment for small perturbation strength $\delta$ is given by \cref{eq:alignement_prediction}. Different panels show results for different aux-head sizes. We observe that the theoretical upper-bound matches well for the first hidden layer of the network, but the second layers alignment breaks away more quickly then expected. We speculate this is due to the additional noise term from the latent-representation in \eqref{eq:auxpertgrad}, having a more direct influence on the second layer, as its weights directly affect the latent-space representation.}
    \label{fig:subllt}
\end{figure}

\subsubsection{Trainable aux head and stability during training}
\label{sec:aux_head_stability}
So far we have only considered the beginning of training and ignored any potential changes the aux head undergoes during training of the student - although in the typical setup the aux head is often chosen to be trainable. For this reason we investigate the behavior of the aux head during training and show that a) the student aux head can surprisingly recover even if initialized randomly and re-align with the teacher during training, giving additional stability to the representation learning and b) this "stability" of the aux head can fail in (very) high-dimensional latent-spaces when noisy gradients induce random drift.

To study the behavior during training, we again consider the MSE loss for the student. In terms of the row-vectors $\alpha_i$ of the aux head, the loss for noise samples $n^{(i)}$ becomes
\begin{equation}
    \mathcal{L}(\theta^{(S)},\Omega_A^{(S)})=\frac{1}{2}\sum_{i=1}^{N}\sum_{j=1}^m \left(\langle\alpha_j^{(T)},f_{\theta^{(T)}}(n^{(i)})\rangle -  \langle\alpha_j^{(S)},f_{\theta^{(S)}}(n^{(i)})\rangle\right)^2\;.
\end{equation}
If we only consider updates of the aux head, minimizing the loss above becomes a linear regression task, where each component vector of the aux head $\alpha_j^{(S)}$ independently tries to predict the corresponding teacher output on the $j$-th aux-neuron $\langle\alpha_j^{(T)},f_{\theta^{(T)}}(n^{(i)})\rangle$, given the data-vector $f_{\theta^{(S)}}(n^{(i)})$.

Leveraging the similarity between teacher and student, we again use a Taylor approximation of the teacher network evaluated at the student weights, i.e. $f_{\theta^{(T,\text{final})}}(n^{(i)}) \approx f_{\theta^{(S)}}(n^{(i)})+J^{(i)}(\theta^{(T,\text{final})}-\theta^{(S)})$ which results in the following approximation for the loss
\begin{equation*}
    \mathcal{L}(\theta^{(S)},\Omega_A^{(S)}) \approx \frac{1}{2} \sum_{i=1}^N\sum_{j=1}^m \left(\langle \alpha_j^{(T)}, f_{\theta^{(S)}}(n^{(i)})+\underbrace{J^{(i)}(\theta^{(T,\text{final})}-\theta^{(S)})}_{\text{representation difference}}\rangle - \langle \alpha_j^{(S)}, f_{\theta^{(S)}}(n^{(i)})\rangle\right)^2\;.
\end{equation*}
Arguably, the latent representation difference of random noise, between teacher and student does not have any meaningful linear correlation with the student representation $ f_{\theta^{(S)}}(n^{(i)})$ and cannot be predicted with linear regression. Hence this term effectively forms uncorrelated residuals $R_{j}^{(i)}$ and we get an approximate loss of
\begin{equation}
\mathcal{L}(\theta^{(S)},\Omega_A^{(S)}) \approx \frac{1}{2} \sum_{i=1}^N\sum_{j=1}^m \left(\langle \alpha_j^{(T)}-\alpha_j^{(S)}, f_{\theta^{(S)}}(n^{(i)})\rangle + R_{j}^{(i)} \right)^2\;.\label{eq:auxstab}
\end{equation}
From this formulation, it becomes obvious that the loss is minimized for the aux head when $\alpha^{(S)}_j =\alpha_j^{(T)}$ and thus the loss gradient will push $\Omega_A^{(S)}$ towards $\Omega_A^{(T)}$. Meanwhile, improving the representation of the student by updating $\theta^{(S)}$ reduces the residuals $R_j^{(i)}$. Conclusively, if the teacher and student share an initialization, the student aux head will remain close to $\Omega^{(T)}$ during training, preventing any distortion of the representation learning discussed above. Importantly, a perturbed student aux head can also recover during training moving back towards the teacher aux head given the gradient signal from \eqref{eq:auxstab}. However, this is strongly reliant on close teacher and student weights, i.e. having the same initialization $\theta^{(S,\text{init})}=\theta^{(T,\text{init})}$ (and the teacher's weights not changing too much during its training), since the recovery relies on a meaningful remaining correlation of the latent-space representation the trained teacher and the student at the beginning of training.

\textbfplus{Subliminal learning is possible with random aux heads}
We have seen that given $\theta^{(S,\text{init})}=\theta^{(T,\text{init})}$ - and assuming the teacher's weights do not change too much during training - we have a clear gradient signal for the student aux head towards the teacher aux head. A surprising consequence of this is that, in principle, subliminal learning is even possible with a randomly initialized student aux head. For this to work, the student has to first learn the correct aux head and then update its representation. In practice both of course happens at the same time, and therefore one expects a much reduced transfer in performance, given the student will train some of its representation on a strongly misaligned aux head. This of course, depends crucially on the shared initialization of teacher and student. In the absence of this, we do not recover a training signal for the aux head and the student will learn a randomly rotated version of the teacher's representation, failing at subliminal learning.

\label{sec:high_d_failure}
\textbfplus{Aux-head drift in high dimensional latent-space} The previous arguments for stability and recovery of the aux head have another failure mode, beside a non-matching initialization of teacher and student, which is due to noisy gradients in the case of a (very) high dimensional latent space. Evaluating the gradient of \cref{eq:auxstab}, with respect to a student aux-head vector $\alpha_j^{(S)}$ yields
\begin{equation}
    G := - \nabla_{\alpha_j^{(S)}}\mathcal{L} = \sum_{i=1}^N \left(\langle\alpha_j^{(T)}-\alpha_j^{(S)},z^{(i)}\rangle+R_j^{(i)}\right) z^{(i)}\;.
\end{equation}
with $z^{(i)}=f_{\theta^{(S)}}(n^{(i)})$. We analyze this gradient $G$ in the case of a very high dimensional $d \gg 1$ latent-space by decomposing it. Using $\Delta \alpha_j := \alpha^{(T)}_j-\alpha_j^{(S)}$ we decompose the latent-representations of the noise data into $z^{(i)}=z^{(i)}_\parallel+z^{(i)}_\perp$, where $z^{(i)}_\parallel \parallel \Delta \alpha_j$ and $z^{(i)}_\perp \perp \Delta \alpha_j$. This results in a component $G_\parallel$ parallel to $\Delta \alpha_j$ and an orthogonal component $G_\perp$. Let us first analyze the parallel component, that contains the "signal" that is actually desirable for the student's learning the teacher representation and keeps student and teacher aux heads aligned
\begin{equation}
    G_\parallel = \sum_{i=1}^N (\underbrace{<\Delta \alpha, z_\parallel^{(i)}>}_{\text{signal}} + \underbrace{R_j^{(i)}}_{\text{noise}})z_{\parallel}^{(i)}\;.
\end{equation}
Assuming an (approximately) variance-preserving initialization of the network, the signal term should scale as $\mathcal ||\Delta \alpha_j||_2 N$ while the noisy residual term scales as $ \sqrt{\mathbb V(R)N}$, ensuring the signal dominates the loss gradient for large numbers of noise samples $N$ during student training. This is however not always true for the orthogonal gradient component, which does not contain any signal and only noise. We have
\begin{equation}
G_\perp=\sum_{i=1}^N (\langle\Delta \alpha, z_\parallel^{(i)}\rangle + R_j^{(i)}) z_\perp^{(i)}
\end{equation}
whose norm scales as $\sqrt{||\Delta \alpha||_2^2 +\mathbb V(R)} \cdot\sqrt{dN}$.\footnote{$||z_\perp^{(i)}|| \propto \sqrt{d}$ if a variance preserving initialization of the networks is used.} Comparing the size of the "signal" gradient and the orthogonal noise component, it becomes clear that for very large $d$ random noise may persist in the gradient, as it does not average out with many samples as quick as in lower dimensional settings, and causes the aux head to randomly rotate during training, which carries problems for potential distortion of the student's learned representation compared to the teacher. When exactly this break down happens, depends however on training specifics like the optimizer, batch size, the number of noise samples used and on the strength of the residuals $R_j^{(i)}$, which are themselves related to the distance between teacher and student representation.

While this aux-head rotation may negatively impact the performance of subliminal learning for architectures with high dimensional latent representations, as simple fix is to make the aux head non-trainable to prevent random rotations. This "fix" applies at least to the simple controlled setting we study in this work. For more complicated architectures like LLMs there may be no clear distinction between \enquote{aux head} and \enquote{class head} in practice. Though as will be discussed below the noise failure mode for high $d$ occurs in the class head as-well. Therefore a solution for more complex architectures might be to fix the entire output-head during training.

\subsubsection{Representation learning from random initializations}
\label{sec:mb_cross_arch}
So far, we have considered teacher and student to belong to the same model class, and having the same initialization, while investigating what happens if the aux head of both is not the same. Now we will reverse the setting and consider student and teacher to have completely independent initializations, even different model architecture, but we will assume the aux head and class head to be the same for both\footnote{Technically one can relax this setting slightly and assume that the output-heads are similar but not identical and still get statistically a "better than guessing" performance. However, subliminal-learning performance is already strongly reduced in the current setting compared to the same initial weights setting and will quickly degrade the more dissimilar the output-heads become. Hence we focus on the case of the same output heads for conciseness.}. We will now show how subliminal learning can still occur even without any similarity of teacher and student. This argument mainly relies on the insight, that in this general setting the student can still learn to match the teacher's latent representation on a sub-space of the latent-space which can allow for a part of the teacher's classification signal to be recovered from the student.\\

Let the class and aux head be
\begin{align}
    \Omega_A &= \begin{bmatrix}
           \alpha^\intercal_{1} \\
           \vdots \\
           \alpha^\intercal_{m}
\end{bmatrix}  &\text{and}  &&\Omega_c &=\begin{bmatrix}
           \omega^\intercal_{1} \\
           \vdots \\
           \omega^\intercal_{n}
\end{bmatrix}\;,
\end{align}
with $\alpha_{i},\omega_c \in \mathbb R^{d}$. Let $\hat \alpha_i$ and $\hat \omega_c$ refer to the normalized versions of the respective vectors. As before, let $z^{(T)}$ and $z^{(S)}$ be latent-space output of the teacher and student model for an input vector $x$. Define $C := \mathrm{span}(\omega_1,..,\omega_n)$ as a subspace of the latent space spanned by all the class-head vectors. We now decompose the latent space into $\mathbb R^d = C \oplus C^\perp$ and call $P_C$ the projection onto $C$. Any vector component in the orthogonal complement $C^\perp$, has no influence on the logit-outputs of the class head. During training the teacher has learned to separate input classes within the sub-space $C$. Effectively that means that if $x$ belongs to class $c$, the latent representation of the data-point will - on average - be shifted by some quantity $\mu$ in the direction of $\omega_c$, while also possessing an irrelevant residual part $z^{\perp_C}$ in the complement of $C$ i.e. 
\begin{equation}
z \approx z_0 + \mu \hat \omega_c + z^{\perp_C}\;,
\label{eq:signal}
\end{equation} 
which results in the logit value for $c$ being $\approx \mu$ larger than the other class logits. 

We now explain how the student can recover part of this representation. Let $A := \mathrm{span}(\alpha_1,..,\alpha_m)$ be the sub-space that has influence on the output of the aux head and let $P_A$ be the orthogonal projection onto this sub-space. During training on noise samples, the student can learn to match the teacher arbitrarily well on the subspace $A$, that is
\begin{equation}
P_A f^{(S)}(x) = P_A f^{(T)}(x)\;.
\end{equation}
Because of the randomly chosen aux head, $P_A$ acts as a random projection of $\mathbb R^d$ onto an $m$-dimensional subspace. The teacher's classification signal from \eqref{eq:signal} gets projected onto $A$ as
\begin{equation}
    P_A z = P_A z_0 + \mu P_A\hat \omega + P_A z^{\perp_C}\;.
    \label{eq:studentmatchC}
\end{equation}
Accordingly, data points of each class are shifted systematically in directions of the projected class vectors $P_A \hat \omega_c$. Therefore the student can still learn to (statistically) separate points of different classes on the sub-space $A$. Unfortunately, also the noise / unrelated representation part $\hat z^{\perp_C}$ gets projected into $A$, so this signal gets weakened by additional noise. An additional complication is, that in the high dimensional latent-space $d \gg n$, the class vectors $\omega_c$ are approximately orthogonal (see \cref{sec:mb_class_head_stability}), allowing for simple distinction of the different classes. In $A$, the respective projections $P_A \hat \omega_c$ may become significantly more collinear, especially if $m < n$, reducing discriminative performance between different classes. After the student has learned to match the teacher and is probed on real data, the final representation vector looks like this
\begin{equation}
    f^{(S)}(x)= P_Af^{(T)}(x) + z^{\perp_A}
\end{equation}
with residual untrained noise in the complement of $A$ that never got any gradient signal to update its representation. Evaluating this embedding on the aux head, gives the following logit:
\begin{equation}
    \langle \omega_{c'}, f^{S}(x)\rangle= \underbrace{\mu \langle \omega_{c'}, P_A\hat \omega_c \rangle}_{\text{signal}} + \underbrace{\langle \omega_{c'},P_A z^{\perp_C} + z^{\perp_A}\rangle}_{\text{noise}}\;.
\end{equation}
Because $\langle \omega_c,P_A\hat \omega_c\rangle > 0$ we can always recover some of the teacher's original output signal, and thus the student should be able to make better predictions than random chance. In practice, the student performance is limited by two factors: Through the "double" projections - once onto the aux-head space and then the class-head space, we get two additional noise components from the respective complements. At the same time the signal strength and specificity can be strongly diminished. As a random projection of dimension $m$, $P_A$ captures a fraction $\frac{m}{d}$ of the total variance in the latent space, and thus $<\hat \omega_c, P_A \hat \omega_c> \approx \frac{m}{d}$, reducing the signal strength by this factor compared to the teacher. Additionally, some class vectors might become more collinear through projection, reducing signal specificity. Therefore the performance of subliminal learning in this setting depends mostly on the number of aux heads $m$. For good performance one needs $m > n$ and $m$ to be a reasonable fraction of $d$. 

One failure mode of subliminal learning in this setting, that has gone unaddressed so far is the student's capability to match the teacher on $C$ assumed in \eqref{eq:studentmatchC}. In a cross architecture setting, if the student network is too simple, it might fail to reproduce the teacher's more complex map. Somewhat counterintuitively, the opposite can also be true: If the student network is too expressive, it might overfit on the teacher's noise samples and fail to generalize them out of distribution. One therefore has to match teacher and student model reasonably in terms of expressiveness for subliminal learning to occur.

\subsubsection{Conclusion on representation learning condition}
\label{sec:whatifnothingisthesame}
To conclude on the previous section, we summarize that for the student to being able to learn the teacher's latent-space representation one requires at minimum either
\begin{itemize}
    \item student and teacher have the same  (or very similar) initialization $\theta^{(S,\text{init})} \approx \theta^{(T,\text{init})}$
    \item both networks share the same initial aux head $\Omega_A^{(S,\text{init})}\approx\Omega_A^{(T,\text{init})}$ and if teacher and student have a different architecture, then both need to be similar in expressiveness.
\end{itemize}
While having either of these conditions allows for some learning of the teacher's representation, resulting in better than random student classification accuracy, good student performance on the classification task requires both conditions to be satisfied. If neither condition is satisfied the student will not be able to learn the teacher's latent-representation.

\subsection{Class Head Condition}
\label{sec:class-head}
Aside from being able to learn the teacher's latent-representation, for the student to inherit the teacher's classification performance the class head of the student and teacher have to match as-well, that is one requires
\begin{equation}
\Omega_C^{(T,\text{final})} \approx \Omega_C^{(S,\text{init})}\label{eq:thlateclasshead}\;.\end{equation}
While the student class head never changes during training, the teacher's class head does and so $\Omega_C^{(T,\text{init})} \approx \Omega_C^{(S,\text{init})}$ is in general technically not sufficient to guarantee student performance.

Below, we first establish that indeed student and class head need to be related for the student to achieve any meaningful classification accuracy, by proving that subliminal learning will not work for a random class head. To determine when \eqref{eq:thlateclasshead} is satisfied, we first discuss why (in most cases) the teacher class head will not change strongly during training, s.t. $\Omega_C^{(T,\text{init})} \approx \Omega_C^{(S,\text{init})}$ in practice does suffice as class-head condition. We then explore how perturbations of the class-head effects student performance to determine how close teacher and student class head have to be for subliminal learning to work. 

\subsubsection{Subliminal learning with random class heads is impossible}
\label{sec:random_class_head_impossible}
We will briefly prove that if teacher and student do not share the same class heads and instead possess independent randomly initialized weights $\Omega_C^{(T,\text{init)}}$ and $\Omega_C^{(S,\text{init})}$, subliminal learning is impossible.

Let $f_{\theta^{(S)}}$ be the student network after training. Even if $f_{\theta^{(S)}}$ does approximate the teacher's output in the latent-space arbitrarily well, initializing the student class head at random, will on average never yield a better classification performance than guessing. This is because, as established above, the student class head is never trained and remains in its initial state. Let 
\begin{equation}
    (\Omega_C^{(S,\text{init})},b_C^{(S,\text{init})}) \sim \mathcal{P}
\end{equation}
be the probability distribution from which the initial values for the student class head are drawn. We presume this distribution to be permutation invariant, i.e. for any permutation matrix $\Pi$ it is equally likely to draw $(\Pi \Omega_C,\Pi b_C)$ as it is to draw $(\Omega_C, b_C)$. This is the case of all typical weight initialization methods that sample each weight value independently. Given a test dataset $x^{(1)},..,x^{(N)}$ with class labels $c_i$ we define the classification accuracy of the student to be 
\begin{equation}
    a = \frac{1}{N}\sum_{i=1}^{N} \mathbbm{1}\left(c_i = \mathrm{argmax}_{c=1..n}\left(\Omega_C^{(S,\text{init})} f_{\theta^{(S)}}(x^{(i)})+b_C^{(S,\text{init})}\right)\right)
\end{equation}
where the networks assigned prediction is the class predicted as most likely. The average accuracy, given any trained student evaluates to $\frac{1}{n}$ through trivial computation:\begin{equation}
\begin{split}\mathbb E(a|f_{\theta^{(S)}})=\int \frac{1}{N}\sum_{i=1}^{N} \mathbbm{1}\left(c_i = \mathrm{argmax}_{c=1..n}\left(\Omega_C f_{\theta^{(S)}}(x^{(i)})+b_C\right)\right) d\mathcal{P}(\Omega_C,b_C)\\
=\frac{1}{N}\sum_{i=1}^{N} \frac{1}{n!}\int \sum_{\Pi \in S_n} \mathbbm{1}\left(c_i = \mathrm{argmax}_{c=1..n}\left(\Pi \Omega_C f_{\theta^{(S)}}(x^{(i)})+\Pi b_C\right)\right) d\mathcal{P}(\Omega_C,b_C)\\
= \frac{1}{N}\sum_{i=1}^{N} \frac{1}{n!}\int (n-1)! d\mathcal{P}(\Omega_C,b_C)=\frac{1}{n}\;.
\end{split}
\end{equation}
However, it is noteworthy that, even with a random class head, individual realizations of a student may outperform guessing significantly: Assume a toy scenario in which a teacher has learned to perfectly separate each input class into $n$ singular points in the latent-space. Suppose the student has inherited this representation through the training on the aux-signal, evaluating the student with a random class head will essentially produce a random permutation $\Pi \in S_n$ of the true class labels. If the permutation has $k$ fix-points, the accuracy will be $\approx \frac{k}{n}$. It is hence not uncommon for a specific student with random class head to show accuracies above $20\%$ when evaluated on MNIST, even though the average student never beats $10 \%$.

\subsubsection{Class head stability during training}
\label{sec:mb_class_head_stability}
For subliminal learning to work, the class head of the teacher cannot change too much during training, given that the student is left with its initial class head. Therefore it might be somewhat surprising, that subliminal learning works well for MLPs when starting from a random weight initialization. Thus we discuss why the teacher class head stays aligned during training, allowing for the the initial and final class head remain compatible. For convenience again write the class head again in terms of its row-vectors $\omega_c \in \mathbb R^d$
\begin{align}\Omega_C &=  \begin{bmatrix}
           \omega^\intercal_{1} \\
           \vdots \\
           \omega^\intercal_{n}
\end{bmatrix}\;,\end{align}
and we define $z^{(i)}:=f_{\theta^{(T)}}(x^{(i)})$ as the teacher's latent-representation of the training data $x^{(i)}$ with associated class labels $c_i$. With this in place, the teacher's loss during training can be written as
\begin{align}
    \mathcal{L}(\theta^{(T)},\Omega_C^{(T)},b_C^{(T)}) &= -\sum_{i=1}^N \log{\left(p^{(T)}(c=c_i|z^{(i)})\right)}\\ &= -\sum_{i=1}^N \left(\langle \omega_{c_i},z^{(i)}\rangle + b_{c_i}-\log\left(\sum_{c=1}^n \exp\left(\langle\omega_c,z^{(i)}\rangle + b_c\right)\right)\right)\;.
\end{align}
This results in the following gradients for the class-head vectors and biases
\begin{align}
    -\nabla_{\omega_c}\mathcal{L} &= \sum_{i=1}^N \left(\delta_{c,c_i}  - p_c(z^{(i)})\right) z^{(i)}\\
    -\frac{\partial \mathcal L}{\partial b_c} &= \sum_{i=1}^{N} \delta_{c,c_i}-p_c(z^{(i)})\;.
\end{align}
We observe that during training the heads bias will change in a manner to equalize the average-predicted class probability and the actual prevalence of the class in the training data. We will assume small initial values for the biases\footnote{As for many initialization procedures, in this work biases are initially sampled from $U(\frac{-1}{\sqrt{d}},\frac{1}{\sqrt{d}})$} and a balanced training set, and ignore bias contributions going forward. Meanwhile the class-head vectors change in the direction of a weighted mean of all data representations $z^{(i)}$ that belong to their class, where sample weights increase the lower their current model assigned probability. At the same time the class head also moves in the opposite direction as the weighted mean of all $z^{(i)}$ that do not belong to its class, where samples with high misclassification probability have increased weight.

\begin{figure}[ht]
    \centering
    \includegraphics[width=0.67\textwidth]{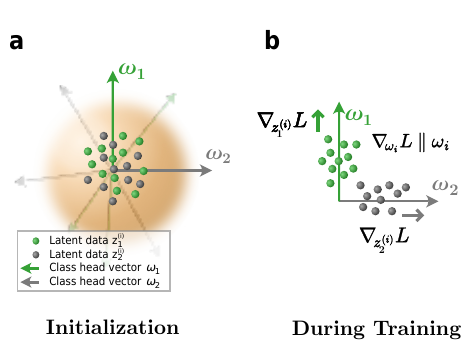}
    \caption{\textbf{The class head remains stable during initial training of a randomly initialized teacher.} \textbf{(a)} After random weight initialization, class-head vectors $\omega_c$ are statistically approximately orthogonal, while the latent-representation of the training data forms an unstructured point-cloud with inseparable classes. \textbf{(b)} After some training symmetry is broken and the representation adapts in the directions of their respective class vectors, which keep their directions through a stabilizing feedback loop.}
    \label{fig:latent_space}
\end{figure}
Finally, the loss gradients w.r.t. representations $z^{(i)}$ are
\begin{equation}
    -\nabla_{z^{(i)}}\mathcal L = \sum_{c=1}^n (\delta_{c,c_i}- p_c(z^{(i})) \omega_c\;.
\end{equation}
pushing the representation in the direction of their correct class-head vector and away from all other class-head vectors. These representations are not directly updated during training but rather the parameters $\theta$ in the model are, and the $z^{(i)}$ actually change according to 
\begin{equation}
    \frac{\partial z^{(i)}}{\partial \theta} = \frac{\partial f_\theta(x^{(i)})}{\partial \theta}\;.
\end{equation}
However, the gradients of $\theta$ are determined by the gradients of the $z^{(i)}$ i.e.
\begin{equation}
    \frac{\partial \mathcal L}{\partial \theta} = \sum_{i=1}^N \frac{\partial \mathcal L}{\partial z^{(i)}} \cdot\frac{\partial z^{(i)}}{\partial \theta}\;.
\end{equation}
Assuming the "black box" network model $f_\theta$ is expressive enough, the representations $z^{(i)}$ will be - on average - updated in the direction of their (negative) loss gradients $ -\nabla_{z^{(i)}}\mathcal L$. This induces the following dynamics of the weights during training, as indicted in \cref{fig:latent_space}: Assuming a high dimensional latent space $d \gg n$, the initial class-head vectors $\omega_c$ are approximately pair-wise orthogonal. The initial latent-representations $z^{(i)}$ are transformations of the input data, by a complex non-linear random map $f_\theta$. Therefore we expect no initial structure of the $z^{(i)}$, that essentially form a random point cloud in the latent space, without any separation of different classes. In this initial configuration, the class heads $\omega_c$ have no strong gradient signal, because the distribution of in-class and out-class vectors in the latent-space is not significantly different (see \cref{fig:latent_space}a). However, the latent-representation has a clear gradient signal: Each point has to move in the direction of its class head and away from the wrong class heads. Thus. at the begin of training, the representation begins to separate along the class-head directions (see \cref{fig:latent_space}b) while the class heads do not significantly change orientation. After this separation has occurred, the class-head vectors get a relevant gradient signal as well. This however, mainly points parallel to their own direction so that the class heads may change scale but do not rotate significantly which results in the different class-head vectors staying approximately aligned with their initial directions after training $\omega_c^{(T,\text{init})}||\omega_c^{(T,\text{final})}$. This ensures that the class-head output logits from the learned teacher's representation are preserved for the student, accordingly $\Omega_C^{(T,\text{init})}$ and $\Omega_C^{(T,\text{final})}$ remain compatible. Similar dynamics of the latent-space representations and class-head are explored the literature in \cite{neural_collapse}.

\begin{figure}[ht]
    \centering
    \includegraphics[width=0.45\textwidth]{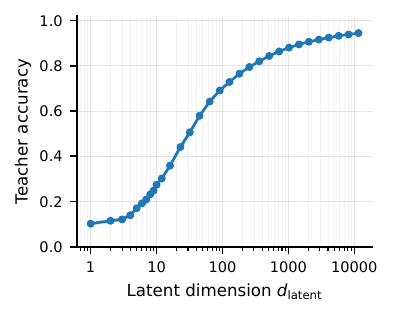}
    \caption{\textbf{Increasing $d_{\mathrm{latent}}$ amplifies the effective supervised signal at the class head by making the frozen latent representation more linearly separable.} Teacher accuracy with only the class head trainable, plotted against latent dimension $d_{\mathrm{latent}}$. Since all earlier layers remain fixed at initialization, performance improvements must arise from the readout alone. Larger latent spaces therefore provide a more favorable random feature basis for linear classification. Mean with $95\%$ CI bootstrap over $20$ seeds.}
    \label{fig:teacher_head_only_d_sweep}
\end{figure}

There are two important caveats to the previous argumentation. If we do not have $d \gg n$, i.e. the number of classes comparable to the latent-space dimension, the random initial class vectors might be somewhat collinear and start to separate to provide better classification signals. Thus one expects more rotation during training and less compatibility in small dimensions $d$. For very large $d$ another concern dominates: Similar to  the previous section \cref{sec:high_d_failure} for the aux-head vectors, in the case of very large latent-dimensions, noise in the gradients may provide another reason for rotation of the class-head and subsequent incompatibility. This is aided by the fact that the data-might become linearly separable in very high dimensions (see  \cref{fig:teacher_head_only_d_sweep}). We therefore expect subliminal learning to work best in the sweet-spot of high $d \gg n$ but not extremely high i.e. $d > 10^3$ dimensions. Though the precise failure dimension depends on many factors like data-set, number of classes and auxiliary outputs and the training procedure.

\subsubsection{Student performance after class-head perturbation}
\label{sec:TheoryClassPerturb}
To assess how a student's accuracy on the classification task is impacted by a mismatched class head, i.e. $\Omega_C^{(S,\text{init})} \neq \Omega_C^{(T,\text{final})}$, we consider a random perturbation with iid Gaussian noise $\epsilon_{ck} \sim N(0,\delta^2)$ with strength $\delta$
\begin{equation}
    \Omega_C^{(S,\text{init})}=\Omega_C^{(T,\text{final})}+\epsilon\;.
\end{equation}
In practice such a perturbation could result from different initial class heads or from class-head changes of the teacher during training. We want to get a rough estimate how the networks performance is impacted by perturbations of strength $\delta$.

Let $l_c = \langle \omega_c, z^{(i)}\rangle$ be the un-perturbed logit belonging to class $c$. With the random class-head perturbation, the logit receives additional additive noise $\langle \epsilon_c,z^{(i)} \rangle$. Let $V$ be the mean-squared size of a component of the data latent-space representations. Then, the perturbed logit approximately follows distribution $l^{(\delta)}_c \sim N(l_c,dV\delta^2)$. Now that we have estimated, how a perturbed class head will impact the logit values, it remains to determine how these perturbed logit signals translate to the actual performance metric: test accuracy.

Suppose we have a network $f_\theta$, that has a classification accuracy of $1-\alpha$, i.e. an error rate of $\alpha$. For an input $x^{(i)}$ of class $c_i$ to be classified correctly, we require the corresponding logit $l_{c_i}>l_{c} \forall c\neq c_i$. Say w.l.o.g the correct class is $c_i=1$. Then for a random test data point we require
\begin{equation}
    \mathbb P(l_1 > \max_{c=2..n}l_c) = 1 - \alpha\;.
    \label{eq:failureprob}
\end{equation}
We are interested in the effect of small perturbations strength $\delta$. In this scenario, the main failure mode of reduced classification performance through perturbation is the case of the "next best class", say w.l.o.g. $c=2$ surpassing the correct class due to the perturbation, i.e. $l_1>l_2$ but $l_2^{(\delta)}>l_1^{(\delta)}$. We therefore consider the simplified case of only two classes and ignore the other ones with lower confusion chance. In this setting \eqref{eq:failureprob} simplifies to $\mathbb P(\Delta l >0)=1-\alpha$ with $\Delta l = l_1-l_2$ and the accuracy of the perturbed network becomes
\begin{equation}
    a(\delta) = \mathbb P(\Delta l^{\delta} >0)
\end{equation}
To estimate $a(\delta)$, we need one additional information: How "certain" the unperturbed network is in its own predictions, which depends on the calibration of the model. By being trained on the Cross-Entropy loss the $\mathrm{softmax}$ output of the network becomes an estimator for the probability $\mathbb P (c=c_i|x)$. In practice, modern neural networks, especially more complex models, tend to be overconfident in their own assessments, though smaller models are at least roughly balanced \cite{calibration}. If the networks own prediction certainty roughly matches its accuracy, the typical difference of the best logit and the next best guess should be roughly
\begin{equation}
\mathbb E(\Delta l )= s\log\left(\frac{1}{\alpha}-1\right)\;.
\label{eq:logitsize}
\end{equation}
Here $s$ is an unknown parameter (effectively a temperature scaling) that indicates whether the network is balanced $s=1$, under-confident $s<1$ or over-confident $s>1$ in its predictions. Given a random data-point of a given class, $\Delta l$ is effectively a random variable, which we model as Gaussian\footnote{This is a rough approximation. In fact, the latent-space components are not necessarily independent.}, given the logits compute as a sum of many different contributions. The two conditions \eqref{eq:failureprob} and \eqref{eq:logitsize}, then enforce
\begin{align}
    \mu&=s\log(1/\alpha -1)\\
    \sigma &= -\mu / q(\alpha)
\end{align}
for $q(\alpha)$ the $\alpha$-quantile of the standard normal distribution. $\Delta l^{(\delta)}$. We now know the perturbed distribution of $\Delta l^{(\delta)}$ and can calculate \begin{equation}
    a(\delta) = \varphi \left(\frac{-q(\alpha)}{\sqrt{1+2dV \left(\frac{q(\alpha)\delta}{s \log{(\alpha^{-1}-1)}}\right)^2}}\right)
    \label{eq:accuracy_scaling}
\end{equation}
with the standard normal cdf $\varphi$.  We can draw the following conclusions from this accuracy formula: The perturbation strength of the class head enters as $\delta^2$, showing robustness of the accuracy at small perturbations. The networks performance is also more resilient at a highly accurate initial network, see  \cref{fig:student_head_perturbation_sweep}f, thanks to the non-linearity connection between of the classification accuracy an the logit-values. This non-linear connection also explains why performance is more robust to changes in the class head compared to the aux head. Precise estimation of the decay in performance given $\delta$ is complicated since the quantities $s$ and $V$ depend on the specific state of the network and scale the effect of $\delta$. though, both variables are expected have roughly unit scale for a small MLP trained with a variance-preserving initialization.

\subsubsection{Conclusion on class-head condition}
We conclude that the condition on class-head compatibility $\Omega_C^{(T,\text{final})}\approx \Omega_C^{(S,\text{init})}$ requires first, the student and teacher class head to have similar initializations and second, that the teacher class head does not rotate much during training. We argue that the latter rarely happens in practice as the teacher's class-head row-vectors remain mostly aligned with their initializations during training, though we point out that misalignment of $\Omega_C^{(T,\text{init})}$ and $\Omega_C^{(T,\text{final})}$ may occur in architectures with very low or very high dimensional latent spaces. Finally, we demonstrated that, while subliminal learning works best for $\Omega_C^{(T,\text{final})}=\Omega_C^{(S,\text{init})}$, student performance will be somewhat resilient to small perturbations of the class head, though accuracy decreases to leading order quadratically in the perturbation strength $\delta$.

\clearpage
\section{Additional Plots}

\begin{figure}[ht]
    \centering
    \includegraphics[width=0.35\linewidth,valign=t]{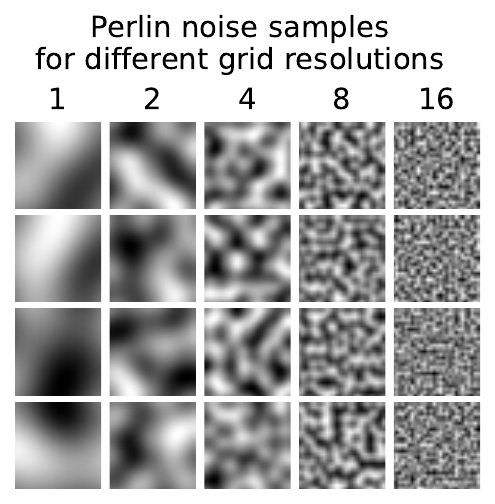}
    \includegraphics[width=0.55\linewidth,valign=t]{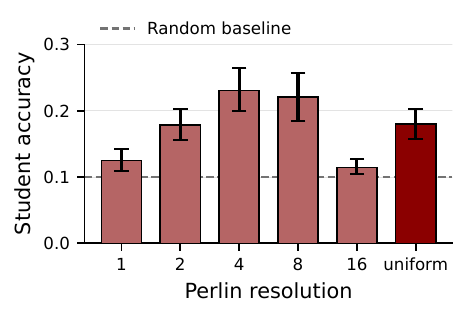}
  \caption{For training a CNN student, we test different resolution levels of spatially correlated Perlin noise \cite{perlin1985image}. This is motivated by data-free distillation work showing that random-noise inputs can induce hidden-layer activation-distribution shift, and by feature-visualization / natural-image-statistics work showing that CNN activations are sensitive to spatial and textural image structure \cite{raikwar2022discovering, olah2017feature, geirhos2018imagenet}. A Perlin resolution of $4$ or $8$ yields higher student accuracy than using uniform noise. }    \label{fig:perlin_noise}
\end{figure}

\begin{figure}[ht]
    \centering
    \begin{subfigure}{0.49\linewidth}
        \centering        
        
        \includegraphics[width=\linewidth]{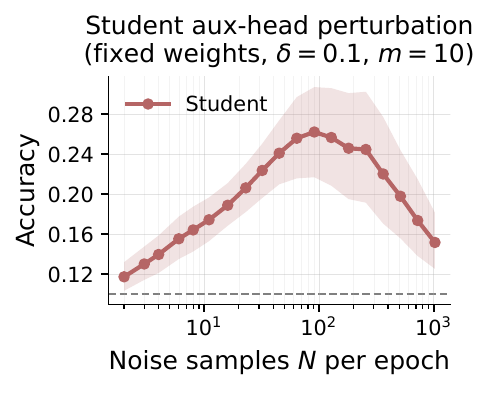}
        \label{fig:noise-steps-aux-p01-sweep}
    \end{subfigure}
    \hfill
    \begin{subfigure}{0.49\linewidth}
        \centering
        \includegraphics[width=\linewidth]{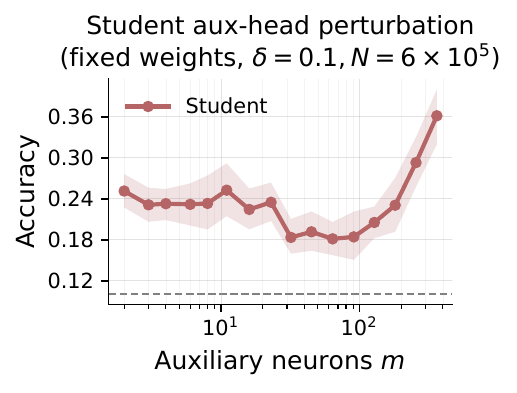}
        \label{fig:m-aux-p01-sweep}
    \end{subfigure}
    \caption{We fix the aux-head weights during training, to separate the effect of self-correction and the effect caused by an increase of $N$ or $m$. Increasing the amount of noise sample per epoch $N$ (left) or auxiliary neurons $m$ (right) cannot arbitrarily compensate for a fixed amount of applied perturbation on the aux head (especially in the $m\ll d=256$ regime, accuracy does not increase as explained in \ref{sec:TheoryPerturbation}.)}
    \label{fig:aux-p01-sweep}
\end{figure}

\begin{figure}
    \centering
    \includegraphics[width=0.9\linewidth]{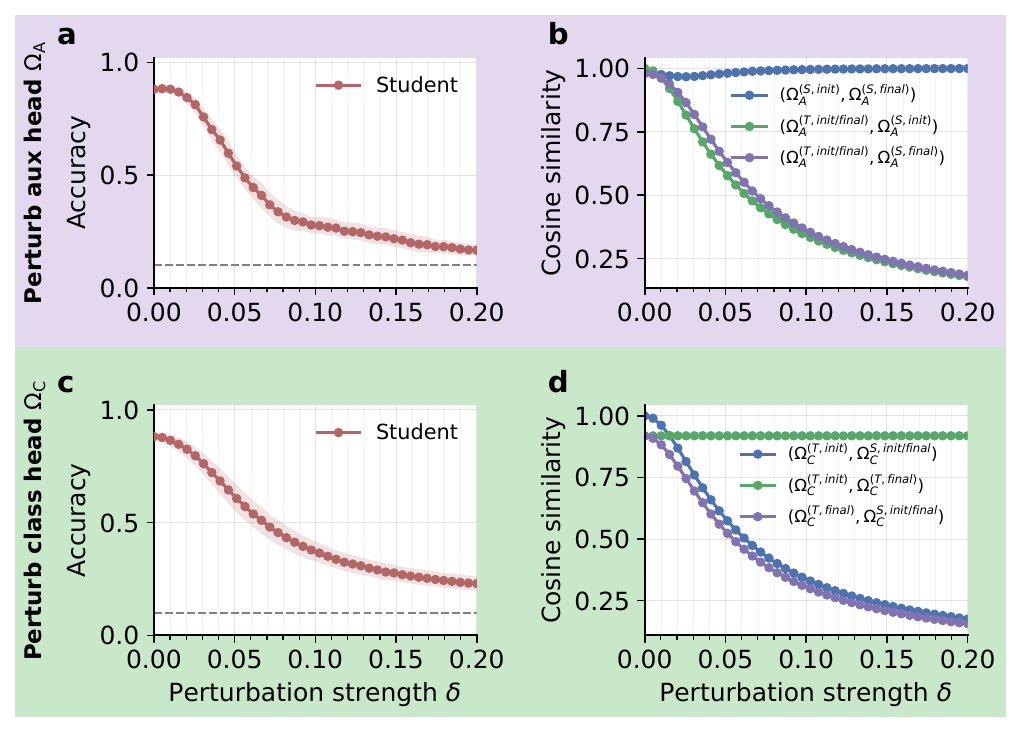}
    \caption{\textbf{Output-head perturbations reveal compatibility limits for subliminal learning.} The scale of the perturbation strength goes beyond the scale of $\approx 0.062$ which is the average weight scale of the network. The decrease in accuracy and cosine similarity observed in \cref{fig:student_head_perturbation_sweep} continues beyond this level.}
    \label{fig:full-perturbation}
\end{figure}

\begin{figure}[ht]
\centering

\includegraphics[width=\linewidth]{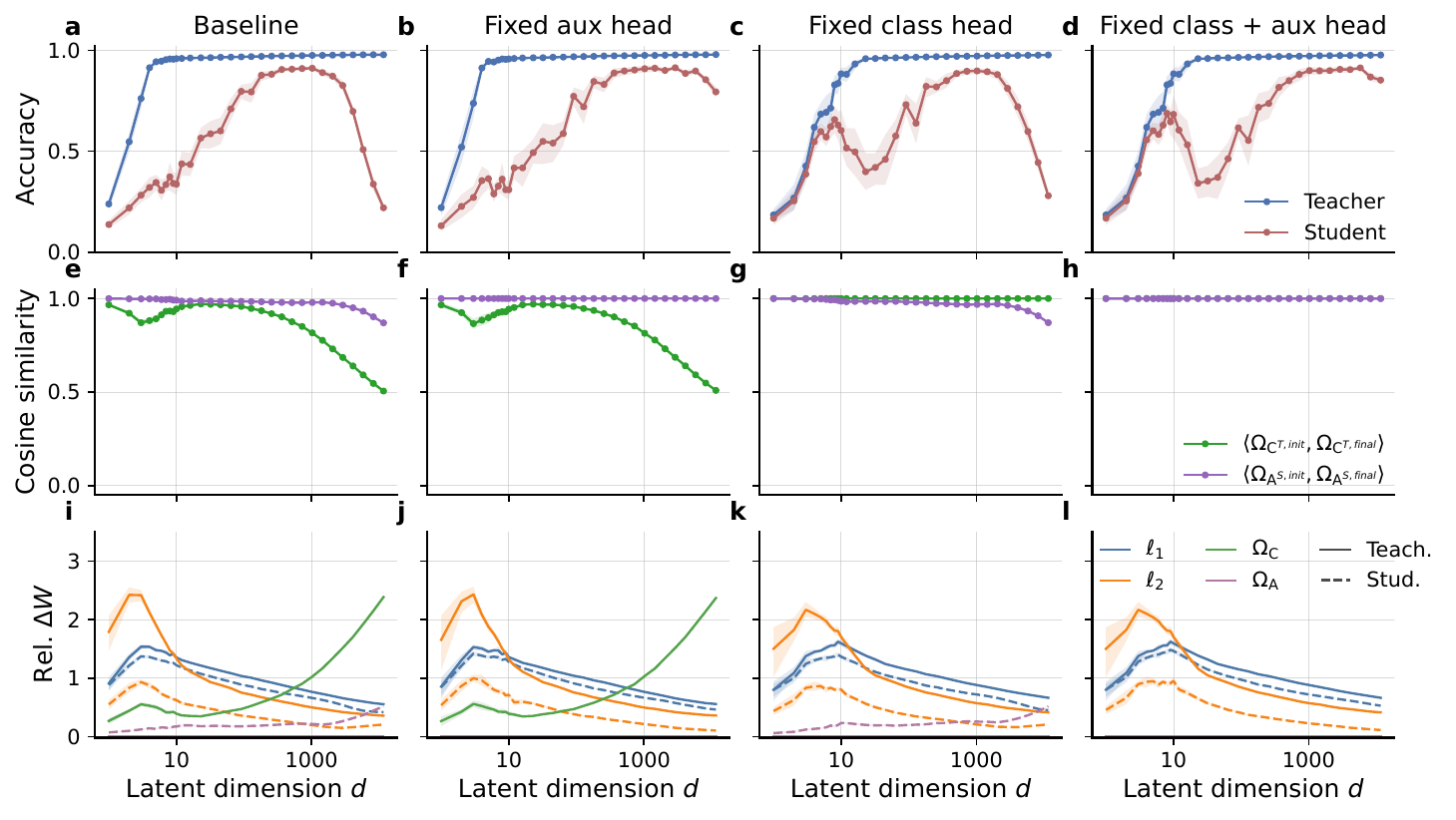}
\caption{\textbf{Fixed-head ablations identify aux-head drift as the dominant high-dimensional failure mode.}
We repeat the latent-dimension sweep while freezing different output heads. 
\textbf{(a,e,i)} In the baseline, student accuracy collapses at large $d$ as both teacher class-head drift and student aux-head drift increase.
\textbf{(b,f,j)} Fixing the aux head strongly reduces the high-$d$ collapse even though the teacher class head still drifts, indicating that class-head drift alone is not the dominant cause.
\textbf{(c,g,k)} Fixing the class head does not prevent the collapse once the aux head drifts, showing that aux-head drift is sufficient to break recoverability.
\textbf{(d,h,l)} Fixing both heads delays and weakens the collapse, but does not remove it entirely, indicating an additional representation-level dimensionality effect.
Values show means over 20 random seeds; shaded regions indicate bootstrapped 95\% confidence intervals.}
\label{fig:d_sweep_ablations}
\end{figure}

\begin{figure}[ht]
    \centering
    \includegraphics[width=1\linewidth]{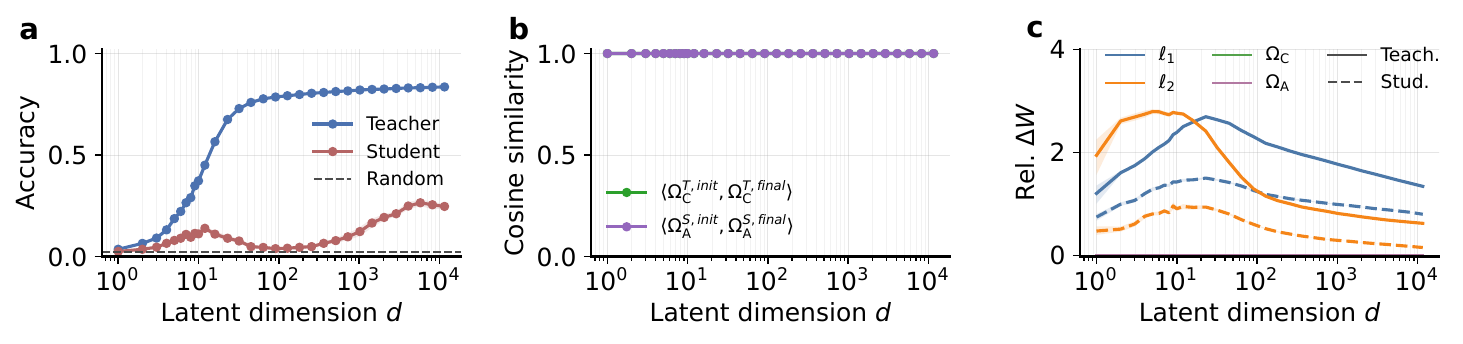}
    \caption{\textbf{The decrease in accuracy also observed for the MNIST case is caused by the size of the aux head.} We repeat the latent dimension sweep performed in \cref{fig:d_sweep_summary} and \cref{fig:d_sweep_ablations} for a setup trained on the balanced EMNIST with $n = 47$ classes. The dip observed in accuracy after $d = m$ indicates that this dip is not caused by the number of classes and rather by the number of auxiliary neurons.}
    \label{fig:d_sweep_emnist}
\end{figure}

\clearpage

\section{Hyperparameters}
\label{apx:experiment_configs}

\begin{table}[ht]
\centering
\caption{
Baseline hyperparameters for the MNIST MLP--MLP subliminal-learning experiments.
The teacher and student use the same split-head architecture and are initialized identically layerwise.
}
\label{tab:baseline_hyperparameters}
\small
\begin{tabular}{ll}
\toprule
\textbf{Category} & \textbf{Value} \\
\midrule
Dataset & MNIST \\
Train / test examples & $60{,}000$ / $10{,}000$ \\
Input preprocessing & \texttt{ToTensor}; normalize with mean $0.1307$, std. $0.3081$ \\
Number of classes & $10$ \\
\midrule
Teacher architecture & MLP with hidden widths $(256,256)$ \\
Student architecture & MLP with hidden widths $(256,256)$ \\
Activation & ReLU after each hidden layer \\
Input dimension & $28 \times 28 = 784$ \\
Shared feature dimension & $d=256$ \\
Classification head & Linear layer $\Omega_C:\mathbb{R}^{256}\to\mathbb{R}^{10}$ \\
Auxiliary head & Linear layer $\Omega_A:\mathbb{R}^{256}\to\mathbb{R}^{10}$ \\
Auxiliary dimension & $m=10$ \\
Parameters per model & $271{,}892$ \\
\midrule
Teacher initialization$^\dagger$ & $(A,A,A,A)$ for $(\ell_1, \ell_2,\Omega_C,\Omega_A)$ \\
Student initialization$^\dagger$ & $(A,A,A,A)$ for $(\ell_1, \ell_2,\Omega_C,\Omega_A)$ \\
Trainable teacher layers & All layers trainable \\
Trainable student layers & All layers trainable \\
Random seeds & $20$ seeds, indexed $0,\ldots,19$ \\
Determinism & Seeded Python, NumPy, PyTorch \\
\midrule
Teacher objective & Cross-entropy on classification logits \\
Student objective & MSE on auxiliary logits \\
Teacher optimizer & Adam \cite{kingma2014adam}, learning rate $10^{-3}$ \\
Student optimizer & Adam, learning rate $10^{-3}$ \\
Adam hyperparameters & PyTorch defaults: $\beta=(0.9,0.999)$, $\epsilon=10^{-8}$ \\
Weight decay & $0$ \\
Teacher epochs & $5$ \\
Student epochs & $5$ \\
Labeled-data batch size & $1024$ \\
\midrule
Noise distribution & Uniform noise, $\mathcal{U}(-1,1)$ independently per pixel \\
Noise normalization & None \\
Noise batch size & $B_{\mathrm{noise}}=1000$ \\
Noise steps per student epoch & $S_{\mathrm{noise}}=60$ \\
Noise samples per student epoch & $N=B_{\mathrm{noise}}S_{\mathrm{noise}}=60{,}000$ \\
Total student noise batches & $5 \times 60 = 300$ \\
Total student noise samples & $300{,}000$ \\
Auxiliary-loss evaluation noise & $10$ batches of size $1000$ \\
\midrule
Perturbations & None in baseline \\
Reported accuracy uncertainty & $95\%$ Bootstrap confidence intervals over 20 seeds \\
\bottomrule
\end{tabular}
\end{table}
$^\dagger$The notation $(A,A,A,A)$ denotes layerwise shared initialization for
$(\ell_1, \ell_2,,\Omega_C,\Omega_A)$.
Although all layers are marked trainable, the teacher aux head receives no
gradient during teacher training because the teacher loss only depends on
classification logits. Similarly, the student class head receives no
gradient during auxiliary training because the student loss only depends on
auxiliary logits.

\clearpage

\begin{table}[ht]
\centering
\caption{
Hyperparameters for the MNIST MLP--CNN cross-architecture subliminal-learning experiment. The teacher uses a three-hidden-layer MLP to approximately match
the CNN student's parameter count. Settings listed in \cref{tab:baseline_hyperparameters} but omitted here are shared.}

\label{tab:mlp_cnn_hyperparameters}
\small
\begin{tabular}{ll}
\toprule
\textbf{Category} & \textbf{Value} \\
\midrule
Teacher architecture & MLP with hidden widths $(256,256,256)$ \\
Student architecture & CNN with channels $(32,128)$ and final feature width $256$ \\
Student conv block 1 & Conv2d$(1,32)$, kernel $3$, padding $1$, ReLU, max-pool $2$ \\
Student conv block 2 & Conv2d$(32,128)$, kernel $3$, padding $1$, ReLU, max-pool $4$ \\
Student final layer & Linear layer $\mathbb{R}^{1152}\to\mathbb{R}^{256}$ with ReLU \\
Shared feature dimension & $d=256$ \\
Classification head & Linear layer $\Omega_C:\mathbb{R}^{256}\to\mathbb{R}^{10}$ \\
Auxiliary head & Linear layer $\Omega_A:\mathbb{R}^{256}\to\mathbb{R}^{10}$ \\
Auxiliary dimension & $m=10$ and $m=50$ \\
Teacher parameters & $337{,}684$ \\
Student parameters & $337{,}620$ \\
\midrule
Teacher initialization$^\ddagger$ & $(R,R,R,A,A)$ for $(\ell_1,\ell_2,\ell_3,\Omega_C,\Omega_A)$ \\
Student initialization$^\ddagger$ & $(R,R,R,A,A)$ for $(\mathrm{conv}_1,\mathrm{conv}_2,\ell_1,\Omega_C,\Omega_A)$ \\
Compatible components & class and aux heads only \\
Random seeds & $20$ seeds, indexed $0,\ldots,19$ \\
\midrule
Noise distribution & Perlin noise with resolution $8$ \\
Noise normalization & None \\
Noise batch size & $B_{\mathrm{noise}}=1000$ \\
Noise steps per student epoch & $S_{\mathrm{noise}}=60$ \\
Teacher epochs & $5$ \\
Student epochs & $5$ \\
\bottomrule
\end{tabular}
\label{tab:hyper_mlp_cnn}
\end{table}
$^\ddagger$Here, $R$ denotes an independently random initialization and $A$ denotes a
shared compatible initialization. Thus, unlike the baseline in ~\cref{tab:baseline_hyperparameters}, the teacher and student do not share feature layers or even architecture; only the output heads are compatible.

\clearpage

\begin{table}[ht]
\centering
\caption{
Hyperparameters for an optimal-condition () MLP--MLP subliminal-learning experiment. Settings listed in \cref{tab:baseline_hyperparameters} but omitted here are shared.
}
\label{tab:perfect_student_hyperparameters}
\small
\begin{tabular}{ll}
\toprule
\textbf{Category} & \textbf{Value} \\
\midrule
Teacher architecture & MLP with hidden widths $(256,256)$ \\
Student architecture & MLP with hidden widths $(256,256)$ \\
Shared feature dimension & $d=256$ \\
Auxiliary dimension & $m=500$ \\
Classification head & Linear layer $\Omega_C:\mathbb{R}^{256}\to\mathbb{R}^{10}$ \\
Auxiliary head & Linear layer $\Omega_A:\mathbb{R}^{256}\to\mathbb{R}^{500}$ \\
\midrule
Teacher initialization & $(A,A,A,A)$ for $(\ell_1,\ell_2,\Omega_C,\Omega_A)$ \\
Student initialization & $(R,R,A,A)$ for $(\ell_1,\ell_2,\Omega_C,\Omega_A)$ \\
Compatible components & class and aux heads \\
Teacher trainable flags$^*$ & $(1,1,0,1)$ for $(\ell_1,\ell_2,\Omega_C,\Omega_A)$ \\
Student trainable flags$^*$  & $(1,1,1,0)$ for $(\ell_1,\ell_2,\Omega_C,\Omega_A)$ \\
Random seeds & $20$ seeds, indexed $0,\ldots,19$ \\
\midrule
Noise distribution & Uniform noise, $\mathcal{U}(-1,1)$ independently per pixel \\
Noise normalization & None \\
Noise batch size & $B_{\mathrm{noise}}=1000$ \\
Noise steps per student epoch & $S_{\mathrm{noise}}=1000$ \\
Noise samples per student epoch & $N=B_{\mathrm{noise}}S_{\mathrm{noise}}=10^6$ \\
Total student noise batches & $5 \times 1000 = 5000$ \\
Total student noise samples & $5 \times 10^6$ \\
\bottomrule
\end{tabular}
\label{tab:optimal_conditions}
\end{table}
$^*$The trainable configurations $(1,1,0,1)$ for the teacher and $(1,1,1,0)$ for
the student freeze the teacher class head and the student aux
head, respectively. Thus, the teacher learns task-relevant feature layers while
keeping a fixed class readout, and the student adapts its randomly
initialized feature layers while keeping the auxiliary readout fixed. Because
the teacher and student share compatible initializations for both output heads,
freezing these heads preserves the auxiliary transfer map and the classification
readout throughout training.
\begin{table}[ht]
\centering
\caption{
Final MNIST test accuracies in an optimal-condition MLP--MLP experiment (\cref{tab:optimal_conditions}). With high joint $m,N$, randomly initialized student feature layers, the student matches
teacher-level performance within bootstrap uncertainty, despite being trained
only on task-unrelated noise.
}
\label{tab:perfect_student_random_init_results}
\small
\begin{tabular}{lcccc}
\toprule
\textbf{Model} & \textbf{Seeds} & \textbf{Mean accuracy} & \textbf{95\% CI} & \textbf{Median / range} \\
\midrule
Teacher & $20$ & $0.96860$ & $[0.96810,\,0.96912]$ & $0.96830$ / $[0.9666,\,0.9711]$ \\
Student & $20$ & $0.96820$ & $[0.96769,\,0.96870]$ & $0.96830$ / $[0.9662,\,0.9704]$ \\
\bottomrule
\end{tabular}
\end{table}

\clearpage
\section{Computational Resources}
No GPUs were used. Experiments were run on a Slurm-managed CPU cluster. The compute nodes used for
our experiments were equipped with 2× Intel Xeon Gold 6130 CPUs at 2.10 GHz,
providing 32 CPU cores per node, with one thread per core,
and 188 GiB of system memory. Each main run (training one student/teacher pair, 5 epochs) used 4 CPU cores
for approximately 2 to 5 minutes. As a conservative upper bound (each run $\le5$min.) for the total compute budget of all conducted experiments we report approximately 3800 CPU-core-hours.

\section{Code Availability}

All code required to reproduce the simulations and analyses presented in this paper is publicly available at \url{https://github.com/Priesemann-Group/subliminal_learning}. The repository includes the simulation scripts for reproducing the results reported in this work.

\end{document}